\documentclass[runningheads,orivec]{./tpl/lncs/llncs}		
\usepackage[pdfsubject={},
            pdftitle={},
            pdfauthor={},
            pdfkeywords={},
            colorlinks=false,
            pdfborder={0.25 0.25 0.25}]{hyperref}
\hypersetup{bookmarksdepth}
\newcommand{\authorhanan}   {Hanan~Alkhammash}  
\newcommand{\authorartem}   {Artem~Polyvyanyy}  
\newcommand{\authoralistair}{Alistair~Moffat}   

\newcommand{\affiliationunimelb}   {The University of Melbourne, Victoria 3010, Australia}                          

\newcommand{\orcidhanan}   {\href{https://orcid.org/0000-0001-5761-1345}{\protect\includegraphics[scale=0.05]{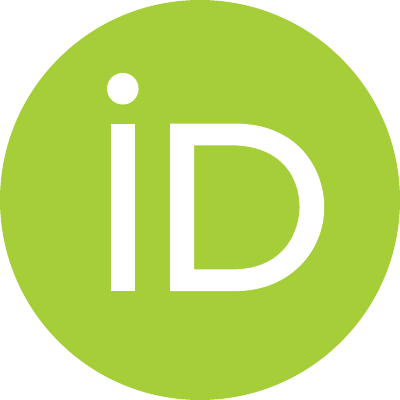}}} 
\newcommand{\orcidartem}   {\href{https://orcid.org/0000-0002-7672-1643}{\protect\includegraphics[scale=0.05]{fig/orcid}}} 
\newcommand{\orcidalistair}{\href{https://orcid.org/0000-0002-6638-0232}{\protect\includegraphics[scale=0.05]{fig/orcid}}} 

\newcommand{\correspondingauthor}{\textsuperscript{(\Letter)}}

\newcommand{\actions}			{{\ensuremath \Lambda}}

\newcommand{\ie}					{i.e.,~}

\newcommand{\articletitle}         {Stochastic Directly-Follows Process Discovery Using Grammatical Inference}
\newcommand{\articletitlerunning}  {\articletitle}
\newcommand{\articleauthor}        {\authorhanan~\inst{1}~\correspondingauthor~\orcidhanan \and \authorartem\inst{1}~\orcidartem \and \authoralistair\inst{1}~\orcidalistair}
\newcommand{\articleauthorrunning}	{\authorhanan \and \authorartem \and \authoralistair}
\newcommand{\articleaffiliation}   {\affiliationunimelb\\\href{mailto:halkhammash@student.unimelb.edu.au}{halkhammash@student.unimelb.edu.au}\\\href{mailto:artem.polyvyanyy@unimelb.edu.au;ammoffat@unimelb.edu.au}{\{artem.polyvyanyy;ammoffat\}@unimelb.edu.au}}

\newcommand{\articlekeyword}       {Process mining, stochastic process discovery, directly-follows graphs}

\newif\ifshowtodos
\showtodostrue		

\usepackage{marvosym}
\usepackage{booktabs}
\usepackage{multirow}
\usepackage{nicefrac}
\usepackage{ifthen}
\usepackage[usenames,dvipsnames]{xcolor}
\usepackage[algoruled,vlined,linesnumbered]{algorithm2e}
\usepackage{amsmath}
\usepackage{mdframed}
\usepackage{paralist}
\setdefaultleftmargin{1em}{1em}{1em}{1em}{1em}{1em}
\usepackage{datetime}
\newdateformat{monthyeardate}{\monthname[\THEMONTH] \THEYEAR}
\hypersetup{bookmarksdepth}
\usepackage[algoruled]{algorithm2e}
\ifshowtodos
\usepackage[textsize=scriptsize,colorinlistoftodos]{todonotes}
\else
\usepackage[textsize=scriptsize,colorinlistoftodos,disable]{todonotes} 
\fi
\let\todonote\todo
\renewcommand{\todo}[2]{\todonote[inline,color=red!20]{TODO (#1): #2}}

\usepackage{tikz,pgf}
\usetikzlibrary{angles,arrows,arrows.meta,petri,backgrounds,automata,trees,plotmarks,shadows,shapes,positioning}
\usepackage{pgfplots}
\pgfplotsset{compat=1.5}
\usetikzlibrary{pgfplots.groupplots}
\usepackage{nicefrac}
\usepackage{url}
\usepackage{mathtools} 
\usepackage{academicons}
\definecolor{orcidlogocol}{HTML}{A6CE39}
\usepackage{multirow}
\usepackage{cleveref}
\usepackage{forest}
\usepackage{oplotsymbl}
\usepackage{colortbl}
\usepackage{rotating}
\usepackage{subfig}
\usepackage{txfonts} 
\usepackage{lipsum}
\usepackage{adjustbox}
\usepackage{csvsimple}
\usepackage{booktabs}
\usepackage{caption} 
\usepackage{wrapfig}

\newcommand{\splitatcommas}[1]{%
  \begingroup
  \ifnum\mathcode`,="8000
  \else
    \begingroup\lccode`~=`, \lowercase{\endgroup
      \edef~{\mathchar\the\mathcode`, \penalty0 \noexpand\hspace{-1pt plus 1em}}%
    }\mathcode`,="8000
  \fi
  #1%
  \endgroup
}

\newtheorem{mytheorem}		{Theorem}
\newtheorem{mydefinition}	{Definition}
\newtheorem{mylemma}			{Lemma}
\newtheorem{myproposition}{Proposition}
\newtheorem{mycorollary}	{Corollary}
\newtheorem{myexample}		{Example}
\newtheorem{myconjecture}	{Conjecture}

\newtheorem{myinvariant}	{Invariant}

\numberwithin{mytheorem}		{section}
\numberwithin{mydefinition}	{section}
\numberwithin{mylemma}			{section}
\numberwithin{myproposition}{section}
\numberwithin{mycorollary}	{section}
\numberwithin{myexample}		{section}
\numberwithin{myconjecture}	{section}
\numberwithin{myremark}			{section}
\numberwithin{myinvariant}	{section}

\newenvironment{define}[3][]
{\begin{mydefinition}[#2]\label{#3}#1\normalfont}
{\hfill\ensuremath{\lrcorner}\end{mydefinition}}

\newenvironment{lem}[3][]
{\begin{mylemma}[#2]\label{#3}#1}
{\hfill\ensuremath{\lrcorner}\end{mylemma}}

\newenvironment{nclem}[3][]
{\begin{mylemma}[#2]\label{#3}#1}
{\end{mylemma}}

\addtocounter{mytheorem}{0}

\newcommand{\mult}{{\ensuremath \,}}

\newcommand{\func}[3]{{{#1}:{#2} \rightarrow {#3}}}

\newcommand{\funcCall}[2]{{\ensuremath {\mathit{#1}}_{\!}\left({#2}\right)}}
\newcommand{\funcCallTwo}[3]{{\ensuremath {\mathit{#1}}_{\!}\left({#2},{#3}\right)}}

\newcommand{\intintervalcc}[2]{{\ensuremath \left[#1 \,..\, #2\right]}}

\newcommand{\intervalcc}[2]{{\ensuremath \left[#1, #2\right]}}

\providecommand{\cardinality}[1]{\ensuremath \left|{#1}\right|}
\renewcommand{\cardinality}[1]{\ensuremath \left|{#1}\right|}

 \newcommand{\mset}[1] {\ensuremath [\splitatcommas{#1}]}

\newcommand{\kleenestar}[1]{{\ensuremath {#1}^{*}}}
\newcommand{\emptysequence}{{\ensuremath \epsilon}}
\newcommand{\sequence}[1]{\ensuremath \langle\splitatcommas{#1}\rangle}

\newcommand{\concat}[2]{\ensuremath #1 \circ #2}

\newcommand{\set}[1]{\ensuremath \{\splitatcommas{#1}\}}
\newcommand{\setbuilder}[2]{\ensuremath \{ #1 \,\mid\, #2 \}}

\newcommand{\pair}[2]{\ensuremath ({#1, #2})}
\newcommand{\triple}[3]{\ensuremath (\splitatcommas{#1,\allowbreak #2,\allowbreak #3})}
\newcommand{\tuple}[1]{\ensuremath (\splitatcommas{#1})}

\providecommand{\holds}{\ensuremath :}
\renewcommand	 {\holds}{\ensuremath :}

\providecommand{\implies}{\ensuremath \Rightarrow}
\renewcommand	 {\implies}{\ensuremath \Rightarrow}

\newcommand{\alphabet}{\mathit{\Sigma}}

\newcommand{\eventLog}[1]{{\mathit{L}_{#1}}}
\newcommand{\trace}[1]{{\mathit{t}_{#1}}}

\newcommand{\length}[1]{{\mathit{|#1|}}}

\newcommand{\multiplicitySeq}[2]{{\funcCall{n}{#1,#2}}}

\newcommand{\action}[1]{\fontfamily{pcr}\selectfont \text{#1}}

\newcommand{\automaton}[1]{\mathit{A}_{#1}}

\newcommand{\probDis}[1]{\mathcal{P}_{#1}}

\definecolor{dfgcolor}{rgb}{0.7,0.13,0.13}
\definecolor{sdfacolor}{rgb}{0.,0.27,0.0}

\definecolor{oldtextcolor}{RGB}{254,157,23}

\definecolor{newtextcolor}{RGB}{130,88,159}

\newcommand{\method}[1]{{\textsl{#1}}}
\newcommand{\ALERGIA}{\method{ALERGIA}}
\newcommand{\DFvM}{\method{DFvM}}

\newcommand{\ENTROPIA}{\method{Entropia}}

\newcommand{\var}[1]{\mbox{\it{#1}}}
\newcommand{\Red}{\var{Red}}
\newcommand{\Blue}{\var{Blue}}
\newcommand{\GASPD}{\method{GASPD}} 

\newcommand{\MDI}{\method{MDI}}
\newcommand{\BMM}{\method{BMM}}
\newcommand{\SM}{\method{SM}}
\newcommand{\IMd}{\method{IMd}}
\newcommand{\IM}{\method{IM}}
\begin{document}
%


\newcommand{\paragraphstart}[2]{{#1}{#2}}

\date{\today}
\title{\articletitle}
\titlerunning{\articletitlerunning}
\author{\articleauthor}
\authorrunning{\articleauthorrunning}
\institute{\articleaffiliation}
\maketitle
\begin{abstract}
Starting with a collection of traces generated by process executions,
process discovery is the task of constructing a simple model that
describes the process, where simplicity is often measured in terms of
model size.
The challenge of process discovery is that the process of interest is
unknown, and that while the input traces constitute positive examples
of process executions, no negative examples are available.
Many commercial tools discover Directly-Follows Graphs, in which
nodes represent the observable actions of the process, and directed
arcs indicate execution order possibilities over the actions.
We propose a new approach for discovering sound Directly-Follows Graphs
that is grounded in grammatical inference over the input traces.
To promote the discovery of small graphs that also describe the
process accurately we design and evaluate a genetic algorithm that
supports the convergence of the inference parameters to the areas
that lead to the discovery of interesting models.
Experiments over real-world datasets confirm that our new approach
can construct smaller models that represent the input traces and
their frequencies more accurately than the state-of-the-art
technique.
Reasoning over the frequencies of encoded traces also becomes
possible, due to the stochastic semantics of the action graphs we
propose, which, for the first time, are interpreted as models that describe the
stochastic languages of action traces.

\end{abstract}

\vspace{-5mm}
\section{Introduction}
\label{sec:intro}
\vspace{-1mm}

Process mining is a discipline that studies data-driven methods and techniques to analyze and optimize processes by leveraging the event data extracted from information systems during process execution.
Process mining approaches can uncover inefficiencies, bottlenecks, and deviations within processes, empowering analysts to make well-informed decisions and formulate hypotheses about future processes~\cite{Aalst2016}.

A fundamental problem studied in process mining is \emph{process
discovery}, which involves constructing process models from event
data~\cite{Aalst2016}.
The discovered models aim to describe the process that generated the
data and can vary in detail and accuracy.
The event data used as input often takes the form of an \emph{event
log}, a collection of \emph{traces}, each captured as a sequence of
executed \emph{actions} in a single instance of the process.
As the same sequence of actions can be executed multiple times by the
process, an event log can contain multiple instances of the same
trace.

A plethora of discovery techniques have been proposed, employing a
range of options to represent the constructed models.
Among these languages, Directly-Follows Graphs (DFGs) stand out for
their intuitiveness, and are a preferred choice for practitioners
seeking insights~\cite{A19CENTERIS,Leemans2019a,ChapelaCampa2022}.
A DFG is a directed graph in which nodes denote actions and arcs
encode ``can follow'' relations between them.
The nodes and arcs of a DFG are annotated with numbers reflecting the
frequencies of the actions and ``occurs next'' dependencies inferred
from the data.

 In this paper, we present an approach grounded in stochastic grammar inference for constructing a Stochastic Directed Action Graph (SDAG), a special type of DFG defined to capture the likelihood of traces, from an event log.
The problem of stochastic grammar inference from a language consists
of learning a grammar representing the strings of the language and
their probabilities, which indicate their importance in the
language~\cite{Higuera2010}.
Hence, discovering a process model from an event log is akin to
grammar inference, where traces of actions in the event log can be
seen as words in the language strings.
There are several reasons why one might choose to use stochastic
inference for process discovery.
First, noisy traces are, in general, infrequent and can thus be
identified and suppressed during inference, noting that noise is
intrinsic to event data and poses challenges to
discovery~\cite{Cheng2015}.
Second, stochastic grammars can be used to predict the next trace or
deduce the probability of the next action given an observed sequence
of actions, information which can inform process
simulations~\cite{Meneghello2023}, decision-making by
analysts~\cite{Aalst2011g}, and future process
design~\cite{Aalst2016}.
Third, grammar inference is performed based on positive example
strings, aiming to: (i) learn the input examples; (ii) favor simpler
explanations of the strings, known as Occam's razor or the parsimony
principle; and to (iii) generalize to \emph{all} positive examples of
the target unknown language~\cite{Higuera2010}.
These aims naturally coincide with the goals of process discovery to
construct models that: (i) are fitting and precise; (ii) simple; and
that (iii) generalize to \emph{all} of the traces the (unknown)
process can support~\cite{Buijs2014}.

We use {\ALERGIA} to perform stochastic grammar inference.
{\ALERGIA} identifies any stochastic regular language from positive
example strings in the limit with probability
one~\cite{Carrasco1994}, with a runtime bounded by a cubic polynomial
in the number of input strings.
In practice the runtime grows only linearly with the size of the
sample set~\cite{Carrasco1994,Higuera2010}.
When performing process discovery, to support process exploration at
different abstraction levels~\cite{Polyvyanyy2008b}, one is often
interested in creating a range of models of various sizes.
In general, the problem of determining whether there is a
representation of a language of a given size is
NP-complete~\cite{Gold1978}.
To control the level of detail in the models it constructs,
{\ALERGIA} makes use of two parameters.
A key contribution in this paper is a genetic algorithm that evolves
an initial random population toward parameter pairs that result in better
models.
Even though SDAGs (unlike DFGs) can have multiple nodes that refer to the same action, we were able to discover SDAGs that are both smaller than the DFGs constructed by a state-of-the-art discovery algorithm and also yield more faithful encodings.

Specifically, we contribute:
\begin{compactenum}
\item 
The first formal semantics of SDAGs (and DFGs) grounded in stochastic languages;
\item
A Genetic Algorithm for Stochastic Process Discovery ({\GASPD}) that discovers a family of SDAGs from an input event log;
\item 
A heuristic for focusing genetic mutations to areas likely to accelerate convergence, resulting in SDAGs of superior quality; and
\item 
An evaluation of {\GASPD} over real-life event logs that both demonstrates its benefits and also suggests future improvements.
\end{compactenum}

\smallskip
\noindent
The remainder of this paper is structured as follows.
The next section discusses related work.
{\Cref{sec:preliminaries}} introduces basic notions required to understand the subsequent sections.
Then, {\Cref{sec:dags}} presents SDAGs and their formal semantics.
{\Cref{sec:spd}} proposes our approach for discovering SDAGs from event logs, while {\Cref{sec:evaluation}} discusses the results of an evaluation of this approach over real-world datasets using our open-source implementation.
The paper concludes with final remarks and discussions in {\Cref{sec:discussion:conclusion}}.
\section{Related Work}
\label{sec:related:work}
{\ALERGIA}, introduced by Carrasco and Oncina~{\cite{Carrasco1994}}, and its variant, {\method{Minimum Divergence Inference}} ({\MDI}) by Thollard et al.~{\cite{TDH00ICML}}, are state-merging algorithms for learning \emph{stochastic deterministic finite automata} (SDFA) from positive examples.
{\MDI} extends {\ALERGIA} with different heuristics and compatibility tests during state merging.
Stolcke and Omohundro~{\cite{SO94CORR}} proposed {\method{Bayesian Model Merging}} ({\BMM}), which deduces a model through structure merging guided by posterior probabilities.

Work by Herbst~{\cite{H00ECML}}, inspired by {\BMM}, was the first application of grammatical inference for stochastic process discovery. 
The approach consists of two routines: model merging and model splitting.
The former generalizes the most specific model by merging processes, using log-likelihood as a heuristic, while the latter refines a general model through iterative splits.
The resulting models are then converted to ADONIS, permitting concurrent behavior.

Recent research in stochastic process mining resulted in several advancements.
Rogge-Solti et al.~{\cite{RAW13BPM}} proposed a technique that lays stochastic performance data over given non-stochastic Petri nets.
Improvements of the algorithm by Burke et al.~{\cite{BLW20ICPM}} introduce five methods to estimate transition probabilities in Petri nets.
An approach developed by Mannhardt et al.~{\cite{MLSL23APN}} discovers data dependencies between Petri net transitions, and Leemans et al.~{\cite{LMS23IS}} extend the approach to capture stochastic long-dependencies triggered by the earlier actions in processes.
To assess the quality of stochastic process models, several quantification techniques have been developed using the Earth Mover's Distance~{\cite{LABP21IS}}, entropy-based conformance checking~{\cite{LP20CAISE}}, and the Minimum Description Length principle~{\cite{Polyvyanyy2020c,AlkhammashPMG22IS}}.

Directly-Follows Graphs emerged as an alternative modeling notation to deterministic automata.
van der Aalst et al.~{\cite{AWM04TKDE}} laid the foundation for process discovery by defining the directly-follows relation within a workflow, capturing the inherent dependencies among activities.
Algorithms like {\method{$\alpha$-Miner}}~\cite{AWM04TKDE}, {\method{Flexible Heuristics Miner}}~\cite{WR11CIDM}, and {\method{Fodina}}~\cite{BW17DSS} discover and map these relations onto Petri nets or BPMN models.

{\method{Directly-Follows visual Miner}} ({\DFvM}) discovers DFGs,
aiming to visually represent the direct dependencies between actions
in the input log.
Designed by Leemans et al.~\cite{Leemans2019a}, {\DFvM} also filters
out less frequent relations, focusing on significant and regular
behaviors.
{\DFvM} consistently constructs high-quality small-sized models
comparable to models produced by top software vendors in the
field~\cite{Polyvyanyy2020c}, and is the method we use as a baseline
in the experiments described in {\Cref{sec:evaluation}}.

Chapela{-}Campa et al.~\cite{ChapelaCampa2022} proposed a Directly-Follows Graphs filter technique to enhance the understandability of DFGs.
The technique formulates the simplification task as an optimization problem, aiming to identify a sound spanning subgraph that minimizes the number of edges while maximizing the sum of edge frequencies.

Widely recognized algorithms like {\method{Inductive Miner}} ({\IM}) by Leemans et al.~\cite{Leemans2013} and {\method{Split Miner}} ({\SM}) by Augusto et al.~\cite{AugustoCDRP18} discover DFGs as an intermediate step, mapping them to well-defined modeling notations.
{\IM} discovers block-structured workflow nets using process trees by computing log cuts based on identified dominant operators such as exclusive choice, sequence, parallel, and loop within the directly-follows graph.
Building upon this approach, {\method{Inductive Miner-directly follows}} ({\IMd})~{\cite{LFA18SOSYM}} handles scalability by employing a single-pass directly-follows graph approach for incomplete logs, and those with infrequent behavior.
{\SM} transforms logs into graphical Directly-Follows Graphs.
The algorithm detects concurrency, prunes the graph, and models it in BPMN, balancing precision and fitness and keeping the model complexity low.

The adaptive metaheuristic framework~{\cite{ADRLB21SOSYM}} optimizes the accuracy of DFG-based process discovery using three strategies that guide exploring the solution space of discoverable DFGs.
The framework iteratively refines DFGs through metaheuristics, evaluates their performance using an objective function, and selects optimal process models.
\vspace{-2mm}
\section{Preliminaries}
\label{sec:preliminaries}
An SDFA is a representation of the traces of a process and their likelihoods.

\begin{define}{Stochastic deterministic finite automata}{def:SDFA}{\quad\\}
A \emph{stochastic deterministic finite automaton} (SDFA) is a tuple $(S, \actions, \delta, p, s_0)$, where
$S$ is a finite set of \emph{states}, 
$\actions$ is a finite set of \emph{actions}, 
$\func{\delta}{S \times \actions}{S}$ is a \emph{transition function}, 
$\func{p}{S \times \actions}{\intervalcc{0}{1}}$ is a \emph{transition probability function}, and
$s_0 \in S$ is the \emph{initial state}, such that
$\forall\, s \in S : ({\sum_{\lambda \in \actions} \funcCall{p}{s,\lambda}} \leq 1.0)$.
\end{define}

\noindent
A {\emph{trace}} is a sequence $t \in \kleenestar{\actions}$.
We use $\emptysequence$ to denote the empty trace.
Given two traces $t_1$ and $t_2$, their concatenation $\concat{t_1}{t_2}$ is obtained by joining $t_1$ and $t_2$ consecutively; 
for example $\concat{\emptysequence}{\emptysequence}=\emptysequence$, $\concat{\emptysequence}{\sequence{\action{b},\action{b}}}=\sequence{\action{b},\action{b}}$, and $\concat{\sequence{\action{b},\action{b}}}{\sequence{\action{b},\action{d}}}=\sequence{\action{b},\action{b},\action{b},\action{d}}$.

An SDFA $A=\tuple{S,\actions,\delta,p,s_0}$ encodes \emph{stochastic language} $L_A$ defined using recursive function $\func{\pi_A}{S\times\kleenestar{\actions}}{\intervalcc{0}{1}}$, {that is,} $\funcCall{L_A}{t} = \funcCall{\pi_A}{s_0,t}$, $t \in \kleenestar{\actions}$, where:
\begin{equation*}\label{eq:sdfa:language}
\begin{split}
\funcCall{\pi_A}{s,\emptysequence} 				& := 1.0 - \sum_{\lambda \in \actions} \funcCall{p}{s,\lambda}, \text{and}\\
\funcCall{\pi_A}{s,\concat{\lambda}{t'}}	& := \funcCall{p}{s,\lambda}\,\mult\,\funcCall{\pi_A}{\funcCall{\delta}{s,\lambda},t'}, \lambda \in \actions, t=\concat{\lambda}{t'}.
\end{split}
\end{equation*}

\noindent
Note that $\funcCall{\pi_A}{s,\emptysequence}$ denotes the probability of terminating a trace in state $s \in S$.
It holds that $\sum_{t \in \kleenestar{\actions}}{\funcCall{L_A}{t}} = 1.0$.
\Cref{fig:SDFA} shows an example SDFA in which
$\funcCall{L_A}{\sequence{\action{a},\action{c},\action{e},\action{c}}} = 0.664$, 
$\funcCall{L_A}{\sequence{\action{a},\action{b},\action{c},\action{e}}} \approx 0.028$, and
$\funcCall{L_A}{\sequence{\action{b},\action{b},\action{b},\action{d}}} = 0$.
The probability associated with a trace is an indication of its importance in the language.\footnote{A common mistake is to confuse this with the probability of the trace being in the language.} 

An \emph{event log} $\eventLog{}$ is a finite multiset of traces, where each trace encodes a sequence of observed and recorded actions executed in the corresponding process.
The multiplicity of $\trace{}$ in $\eventLog{}$, denoted by
$\multiplicitySeq{\trace{}}{\eventLog{}}$, indicates how frequently it has been observed and recorded; we thus have
$\length{\eventLog{}} = \sum_{\trace{} \in \eventLog{}} \multiplicitySeq{\trace{}}{\eventLog{}}$.
The empirical finite support distribution associated with $\eventLog{}$, denoted as $\probDis{\eventLog{}}$, is given by $\funcCall{\probDis{\eventLog{}}}{\trace{}} = \multiplicitySeq{\trace{}}{\eventLog{}}/\length{\eventLog{}}$.
For example, suppose that $\eventLog{}=\mset{\sequence{\action{a},\action{c},\action{e},\action{c}}^{1057},\sequence{\action{a},\action{b},\action{c},\action{e}}^{272},\sequence{\action{b},\action{b},\action{b},\action{d}}^{164}}$ is an event log.
It holds that $\length{\eventLog{}}=1493$, $\funcCall{\probDis{\eventLog{}}}{\sequence{\action{a},\action{c},\action{e},\action{c}}} \approx 0.708$, $\funcCall{\probDis{\eventLog{}}}{\sequence{\action{a},\action{b},\action{c},\action{e}}} \approx 0.182$, and $\funcCall{\probDis{\eventLog{}}}{\sequence{\action{b},\action{b},\action{b},\action{d}}} \approx 0.110$.

{\method{Entropic relevance}} relies on the minimum description length principle to measure the number of bits required to compress a trace in an event log using the structure and information about the relative likelihoods of traces described in a model, for instance, an SDFA~\cite{AlkhammashPMG22IS}.
Models with lower relevance values to a given log are preferred
because they describe the traces and their likelihoods better.
For example, the entropic relevance of SDFA $A$ from {\Cref{fig:SDFA}} to event log $L$ is $3.275$ bits per trace.\footnote{We use the uniform background coding model throughout this work {\cite{AlkhammashPMG22IS}}.}

\begin{figure}[t]
\begin{center}
\begin{minipage}{0.49\textwidth}
\centering
    \begin{tikzpicture}[scale=0.8, transform shape, ->, >=stealth', shorten >=1pt, auto, initial text=, node distance=16mm, on grid, semithick, every state/.style={fill=orange!20, draw, text=black, minimum size=9mm}]
\node[initial,state,label=90:$s_0$]				(s0)				{$0.0$};
\node[state,label=270:$s_1$]					(s1)[below=of s0] 	{$0.0$};
\node[state,label=270:$s_2$]					(s2)[right=of s1] 	{$0.0$};
\node[state,label=90:$s_3$]						(s3)[above=of s2]	{$0.20$};
\node[state,label=90:$s_4$]						(s4)[right=of s3]	{$1.0$};

\path (s0) edge node [right]				{$\action{a}$} node [left, font=\scriptsize] 	{$(1.0)$}  (s1)
      (s1) edge node [above]				{$\action{c}$} node [below, font=\scriptsize] 	{$(0.83)$} (s2)
      (s1) edge [loop left] node [align=center]	{$\action{b}$ \\ \scriptsize $(0.17)$} 				   (s1)
      (s2) edge node [right]				{$\action{e}$} node [left, font=\scriptsize] 	{$(1.0)$}  (s3)
      (s3) edge node [above]				{$\action{c}$} node [below, font=\scriptsize] 	{$(0.80)$} (s4);
\end{tikzpicture}
\caption{An SDFA $A$.}
\label{fig:SDFA}
\end{minipage}
\begin{minipage}{0.49\textwidth}
    \centering
    \begin{tikzpicture}[scale=0.8, transform shape, ->, >=stealth', shorten >=1pt, auto, initial text=, node distance=16mm, on grid, semithick, every state/.style={fill=orange!20, draw, text=black, minimum size=9mm}]
\node[initial,state,label=90:$s_0$]				(s0)						{$0$};
\node[state,label=270:$s_1$]					(s1)[below=of s0] 			{$0$};
\node[state,label=270:$s_2$]					(s2)[right=of s1] 			{$0$};
\node[state,label=90:$s_3$]						(s3)[above=of s2]			{$0$};
\node[state,label=90:$s_4$]						(s4)[right=of s3]			{$1057$};
\node[state,label=270:$s_5$]					(s5)[left=of s1]			{$0$};
\node[state,label=90:$s_6$]						(s6)[above=of s5]			{$0$};
\node[state,label=90:$s_7$]						(s7)[left=of s6]			{$272$};

\path (s0) edge node [right] 					{$\action{a}$} node [left, font=\scriptsize] {$(1329)$} 	(s1)
	  (s1) edge node [above]					{$\action{c}$} node [below, font=\scriptsize] {$(1057)$} 	(s2)
      (s2) edge node [right]					{$\action{e}$} node [left, font=\scriptsize] {$(1057)$} 	(s3)
      (s3) edge node [above]					{$\action{c}$} node [below, font=\scriptsize] {$(1057)$} 	(s4)
      (s1) edge node [above]					{$\action{b}$} node [below, font=\scriptsize] {$(272)$} 	(s5)
      (s5) edge node [right]					{$\action{c}$} node [left, font=\scriptsize] {$(272)$} 		(s6)
      (s6) edge node [above]					{$\action{e}$} node [below, font=\scriptsize] {$(272)$} 	(s7);
\end{tikzpicture}
\caption{A PAT $T$.}
\label{fig:PAT}
\end{minipage}
\end{center}
\vspace{-5mm}
\end{figure}
\section{Stochastic Directed Action Graphs}
\label{sec:dags}

We now describe SDAGs, their stochastic semantics, and their
relationship to DFGs.

\begin{define}{Stochastic directed action graphs}{def:SDAG}
A \emph{stochastic directed action graph} (SDAG) is a tuple $(N,\actions,\beta,\gamma,q,i,o)$, where 
$N$ is a finite set of \emph{nodes}, 
$\actions$ is a finite set of \emph{actions}, 
$\func{\beta}{N}{\actions}$ is a \emph{labeling function},
$\gamma \subseteq  (N \times N) \cup (\set{i} \times N) \cup (N \times \set{o})$ is the \emph{flow relation},
$\func{q}{\gamma}{\intervalcc{0}{1}}$ is a \emph{flow probability function}, and 
$i \not\in N$ and $o \not\in N$ are the \emph{input} node and the \emph{output} node,
such that 
$\forall\, n \in N \cup \set{i} : (\sum_{m \in \setbuilder{k \in N \cup \set{o}}{\pair{n}{k} \in \gamma}}{\funcCall{q}{n,m}}=1)$.
\end{define}

\noindent
An \emph{execution} of an SDAG is a finite sequence of its nodes beginning with $i$ and ending with $o$, such that for every two consecutive nodes $x$ and $y$ in the sequence there is an arc $\pair{x}{y} \in \gamma$.
A \emph{trace} of an SDAG is a sequence of actions such that there is an execution that \emph{confirms} the trace in which the nodes, excluding the input and output nodes, are the actions of the trace in the order they appear.
\Cref{fig:SDAG:1} shows example SDAG $G$. 
The sequence of nodes $\sequence{i,n_1,n_3,n_5,n_4,o}$ is an execution of $G$ that confirms trace $\sequence{\action{a},\action{c},\action{e},\action{c}}$.

An SDAG $G$ encodes stochastic language $L_G$ such that for a trace $t$ of $G$ it holds that $\funcCall{L_G}{t}$ is equal to the sum of probabilities of all the executions of $G$ that confirm $t$, where the probability of an execution is equal to the product of the probabilities, as per function $q$, of all the arcs, as per $\gamma$, defined by all pairs of consecutive nodes in the execution. In addition, if $t \in \actions^*$ is not a trace of $G$, it holds that $\funcCall{L_G}{t}=0$.

Given an SDFA, one can obtain its corresponding SDAG.

\begin{figure}[t]
\vspace{-2mm}
\begin{center}
\begin{minipage}{.32\textwidth}
\begin{tikzpicture}[scale=.70, transform shape, ->, >=stealth', shorten >=1pt, auto, node distance=16mm, on grid, semithick, action/.style={fill=dfgcolor!20, draw, rounded corners, minimum size=9mm}]
\node[action]       (n0) {\begin{tabular}{c}\large{$i$}\\\small{1493.0}\end{tabular}};
\node[action,label=0:  $n_1$] (n1) [below=1.8cm and 1cm of n0] {\begin{tabular}{c}\large{$\action{a}$}\\\small{1493.0}\end{tabular}};
\node[action,label=270:$n_2$] (n2) [below left=1.8cm and 1.07cm of n1] {\begin{tabular}{c}\large{$\action{b}$}\\\small{\,\,305.6\,}\end{tabular}};
\node[action,label=0:  $n_3$] (n3) [below right=1.8cm and 1.07cm of n1] {\begin{tabular}{c}\large{$\action{c}$}\\\small{1493.0}\end{tabular}};
\node[action,label=0:  $n_5$] (n4) [below=1.8cm and 1.07cm of n3] {\begin{tabular}{c}\large{$\action{e}$}\\\small{1493.0}\end{tabular}};
\node[action,label=180:$n_4$] (n5) [below=1.8cm and 1.07cm of n2] {\begin{tabular}{c}\large{$\action{c}$}\\\small{1187.4}\end{tabular}};
\node[action]      (n6) [below left=1.8cm and 1cm of n4] {\begin{tabular}{c}\large{$o$}\\\small{1493.0}\end{tabular}};

\path (n0) edge                       node {\small \begin{tabular}{c}$1.00$\\(1493.0)\end{tabular}} (n1)
      (n1) edge [bend right=31][swap] node[left, pos=0.4] {\small \begin{tabular}{c}$0.17$\\(253.7)\end{tabular}} (n2)
      (n1) edge [bend left=31]        node[right, pos=0.4] {\small \begin{tabular}{c}$0.83$\\(1239.3)\end{tabular}} (n3)
      (n2) edge [loop left]           node [left, pos=0.65]{\small \begin{tabular}{c}$0.17$\\(51.9)\end{tabular}} (n2)
      (n2) edge [above=31]            node {\small \begin{tabular}{c}$0.83$\\(253.7)\end{tabular}} (n3)
      (n3) edge                       node {\small \begin{tabular}{c}$1.00$\\(1493.0)\end{tabular}} (n4)
      (n4) edge [below=31]            node {\small \begin{tabular}{c}$0.80$\\(1187.4)\end{tabular}} (n5)
      (n4) edge [bend left=31]        node [right, pos=0.4]{\small \begin{tabular}{c}$0.20$\\(305.6)\end{tabular}} (n6)
      (n5) edge [bend right=31][swap] node [left, pos=0.4]{\small \begin{tabular}{c}$1.00$\\(1187.4)\end{tabular}} (n6);
\end{tikzpicture}
\caption{SDAG $G$.}
\label{fig:SDAG:1}
\end{minipage}
\begin{minipage}{.32\textwidth}
\begin{tikzpicture}[scale=.70, transform shape, ->, >=stealth', shorten >=1pt, auto, node distance=16mm, on grid, semithick, action/.style={fill=dfgcolor!20, draw, rounded corners, minimum size=9mm}]
\node[action]  (n0) {\begin{tabular}{c}\large{$i$}\\\small{1493.0}\end{tabular}};
\node[action,label=0:  $n_1$]  (n1) [below=1.8cm and 1cm of n0] {\begin{tabular}{c}\large{$\action{a}$}\\\small{1493.0}\end{tabular}};
\node[action,label=250:$n_2$]  (n2) [below left=1.8cm and 1.07cm of n1] {\begin{tabular}{c}\large{$\action{b}$}\\\small{\,\,305.6\,}\end{tabular}};
\node[action,label=0:  $n_3$]  (n3) [below right=1.8cm and 1.07cm of n1] {\begin{tabular}{c}\large{$\action{c}$}\\\small{2478.7}\end{tabular}};
\node[action,label=180:  $n_5$]  (n4) [below left=1.8cm and 1.07cm of n3] {\begin{tabular}{c}\large{$\action{e}$}\\\small{1239.3}\end{tabular}};
\node[action] (n6) [below=1.8cm and 1cm of n4] {\begin{tabular}{c}\large{$o$}\\\small{1493.0}\end{tabular}};

\path (n0) edge                       node {\small \begin{tabular}{c}$1.00$\\(1493.0)\end{tabular}} (n1)
      (n1) edge [bend right=31][swap] node[left, pos=0.4] {\small \begin{tabular}{c}$0.17$\\(253.7)\end{tabular}} (n2)
      (n1) edge [bend left=31]        node[right, pos=0.4] {\small \begin{tabular}{c}$0.83$\\(1239.3)\end{tabular}} (n3)
      (n2) edge [loop left]           node[left, pos=0.65] {\small \begin{tabular}{c}$0.17$\\(51.9)\end{tabular}} (n2)
      (n2) edge [above=31]            node {\small \begin{tabular}{c}$0.83$\\(253.7)\end{tabular}} (n3)
      (n3) edge [bend right=21][swap] node[left, pos=0.6] {\small \begin{tabular}{c}$0.50$\\(1239.3)\end{tabular}} (n4)
      (n4) edge [bend right=21][swap] node[right, pos=0.05] {\small \begin{tabular}{c}$0.80$\\(985.7)\end{tabular}} (n3)
      (n4) edge                       node {\small \begin{tabular}{c}$0.20$\\(253.7)\end{tabular}} (n6)
      (n3) edge [bend left=57]        node[right, pos=0.7] {\small \begin{tabular}{c}$0.50$\\(1239.3)\end{tabular}} (n6);
\end{tikzpicture}
\caption{SDAG $G'$.}
\label{fig:SDAG:2}
\end{minipage}
\begin{minipage}{.32\textwidth}
\begin{tikzpicture}[scale=.70, transform shape, ->, >=stealth', shorten >=1pt, auto, node distance=16mm, on grid, semithick, action/.style={fill=dfgcolor!20, draw, rounded corners, minimum size=9mm}]
\node[action]              (n0) {\begin{tabular}{c}\large{$i$}\\\small{\,\,1493\,\,}\end{tabular}};
\node[action,label=0: $n_1$]               (n1) [below right=1.8cm and 1.07cm of n0] {\begin{tabular}{c}\large{$\action{a}$}\\\small{\,\,1329\,\,}\end{tabular}};
\node[action,label=110: $n_2$]              (n2) [below left=1.8cm and 1.07cm of n0] {\begin{tabular}{c}\large{$\action{b}$}\\\small{\,\,\,\,764\,\,\,\,}\end{tabular}};
\node[action,label=0: $n_3$]               (n3) [below=1.8cm and 1.07cm of n1] {\begin{tabular}{c}\large{$\action{c}$}\\\small{\,\,2386\,\,}\end{tabular}};
\node[action,label=180: $n_5$]             (n4) [below left=1.8cm and 1.07cm of n3] {\begin{tabular}{c}\large{$\action{e}$}\\\small{\,\,1329\,\,}\end{tabular}};
\node[action,label=180: $n_4$]             (n5) [below=1.8cm and 1.07cm of n2] {\begin{tabular}{c}\large{$\action{d}$}\\\small{\,\,\,\,164\,\,\,\,}\end{tabular}};
\node[action]             (n7) [below=1.8cm and 1cm of n4] {\begin{tabular}{c}\large{$o$}\\\small{\,\,1493\,\,}\end{tabular}};

\path (n0) edge [bend left=35]          node {\small (1329)} (n1)
           edge [bend right=35] [swap]  node {\small (164)} (n2)
      (n1) edge                         node[above] {\small (272)} (n2)
      (n1) edge                         node {\small (1057)} (n3)
      (n2) edge                         node[right, pos=0.13] {\small (272)} (n3)
           edge [swap]                  node {\small (164)} (n5)
           edge [loop left]             node {\small (328)} (n2)
      (n3) edge [bend left=51]          node {\small (1057)} (n7)
           edge [bend right=21] [swap]  node[left, pos=0.2] {\small (1329)} (n4)
      (n4) edge [bend right=21] [swap]  node[right, pos=0.16] {\small (1057)} (n3)
           edge                         node {\small (272)} (n7)
      (n5) edge [bend right=51] [swap]	node {\small (164)} (n7);
\end{tikzpicture}
\caption{DFG.}
\label{fig:DAG:DFvM}
\end{minipage}
\end{center}
\vspace{-8mm}
\end{figure}

\begin{define}{SDAG of SDFA}{def:SDFA2SDAG}{\quad\\}
Let $A$ be an SDFA $\tuple{S,\actions,\delta,p,s_0}$.
Then, $\tuple{N,\actions,\beta,\gamma,q,i,o}$, where it holds that:\\
\begin{tabular}{lccl}
\textbf{--} & $N$      & $=$ & $\delta$, $i \not\in N$, $o \not\in N$,  \\
\textbf{--} & $\beta$  & $=$ & $\setbuilder{\pair{\triple{x}{\lambda}{y}}{\lambda}}{\triple{x}{\lambda}{y} \in \delta}$, \\
\textbf{--} & $q$      & $=$ & $\setbuilder{\triple{\triple{x}{\lambda_1}{y}}{\triple{y}{\lambda_2}{z}}{\funcCall{p}{y,\lambda_2}}}{\triple{x}{\lambda_1}{y} \in \delta \,\land\,\triple{y}{\lambda_2}{z} \in \delta} \,\cup\,$ \\
            &          &     & $\setbuilder{\triple{i}{\triple{s_0}{\lambda}{y}}{\funcCall{p}{s_0,\lambda}}}{\triple{s_0}{\lambda}{y} \in \delta} \,\cup\,$ \\
            &          &     & $\setbuilder{\triple{\triple{x}{\lambda}{y}}{o}{1.0-\sum_{\mu \in \actions} \funcCall{p}{y,\mu}}}{\triple{x}{\lambda}{y} \in \delta \,\land\, {\sum_{\mu \in \actions} \funcCall{p}{y,\mu}} < 1.0} \,\cup\,$ \\
						&          &     & $\setbuilder{\triple{i}{o}{1.0-\sum_{\mu \in \actions} \funcCall{p}{s_0,\mu}}}{{\sum_{\mu \in \actions} \funcCall{p}{s_0,\mu}} < 1.0}$, and\\
\textbf{--} & $\gamma$ & $=$ & $\setbuilder{\pair{x}{y}}{\triple{x}{y}{z} \in q \,\land\, z \in \intervalcc{0}{1}}$,
\end{tabular}\\
is the \emph{SDAG of $A$}, denoted by $\funcCall{SDAG}{A}$.
\end{define}

\noindent
If $A$ is an SDFA, then $\funcCall{SDAG}{A}$ is \emph{sound}~\cite{Aalst1997} by construction, as every node of $\funcCall{SDAG}{A}$ is on a directed walk from the input node to the output node while the directed walks in $\funcCall{SDAG}{A}$ define all and only its executions. Hence, one can reach the output node from every node of $\funcCall{SDAG}{A}$ (option to complete property), once a directed walk, and thus the corresponding execution, reaches the output node, it completes (proper completion), and the output node is the only deadlock in $\funcCall{SDAG}{A}$ (no dead actions).

In addition, an SDAG of an SDFA has a special structure; that is, it is deterministic.

\begin{define}{Deterministic SDAGs}{def:det:SDAG}
An SDAG $\tuple{N,\actions,\beta,\gamma,q,i,o}$ is \emph{deterministic} 
if and only if
for every two arcs that start at the same node and lead to two distinct nodes
it holds that the labels of the nodes are different, 
that is, it holds that:
$\forall\, n \in N \cup \set{i} \, \forall\, n_1,n_2 \in N \holds ((\pair{n}{n_1} \in \gamma \,\land\, \pair{n}{n_2} \in \gamma \land n_1 \neq n_2 ) \implies \funcCall{\beta}{n_1} \neq \funcCall{\beta}{n_2}).$
\end{define}

\noindent
In a deterministic SDAG, every distinct arc originating at a node must
lead to a node with a different action.
Consequently, every trace of a deterministic SDAG has only one execution that confirms it, significantly simplifying the computation of the stochastic language of the action graph.
The SDAG of an SDFA is deterministic.

\begin{lem}{Deterministic SDAGs}{lem:det:SDAG}
If $A$ is a SDFA
then $\funcCall{SDAG}{A}$ is deterministic.
\end{lem}

\noindent
\Cref{lem:det:SDAG} holds by construction of \Cref{def:SDFA2SDAG}.
The SDAG $G$ from \Cref{fig:SDAG:1} is an SDAG of SDFA $A$ in \Cref{fig:SDFA}, {\ie} it holds that $G = \funcCall{SDAG}{A}$.
As $G$ is deterministic, it holds that $\funcCall{L_G}{\sequence{\action{a},\action{c},\action{e},\action{c}}} = 0.664$, which is equal to the product of the probabilities on all the arcs of the corresponding execution $\sequence{i,n_1,n_3,n_5,n_4,o}$.

\begin{define}{SFA of SDAG}{def:SDAG2SDFA}
Let $G$ be an SDAG $\tuple{N,\actions,\beta,\gamma,q,i,o}$.
Then, $\tuple{S,\actions,\delta,p,s_0}$, where it holds that
$S = N \cup \set{i}$,
$\delta = \setbuilder{(x,\lambda,y) \in (N \cup \set{i}) \times \actions \times N}{(x,y) \in \gamma \land \lambda = \funcCall{\beta}{y}}$,
$p = \setbuilder{\triple{x}{\lambda}{\sum_{\funcCall{\beta}{y}=\lambda,\, y \in N} \funcCall{q}{x,y}}}{x \in N \cup \set{i} \land \lambda \in \actions \land \exists\, z \in N \holds (x,\lambda,z) \in \delta}$, and
$s_o = i$,
is the \emph{stochastic finite automaton} (SFA) of $G$, denoted by $\funcCall{SFA}{G}$.
\end{define}

\noindent
\Cref{def:SDAG2SDFA} generalizes Definition III.5 from~\cite{Polyvyanyy2020c} to account for the fact that an SDAG can have multiple nodes that refer to the same action.

Note that $\funcCall{SFA}{G}$ is indeed an SDFA if $G$ is deterministic.

\begin{lem}{Determinism}{lem:det}
If $G$ is a deterministic SDAG then
$\funcCall{SFA}{G}$ is an SDFA.
\end{lem}

\noindent
\Cref{lem:det} holds by construction of \Cref{def:SDAG2SDFA}.
The SDAG of an SDFA and the SDFA of a deterministic SDAG have the same stochastic languages.

\begin{nclem}{Equivalence}{thm:equivalence}
Let $A$ be an SDFA and let $G$ be an SDAG.\\ Then, it holds that
$\text{(i) } L_A = L_{\funcCall{SDAG}{A}}$,
and
$\text{(ii) } L_G = L_{\funcCall{SFA}{G}}$.
\hfill\ensuremath{\lrcorner}
\end{nclem}

\noindent
The stochastic language of the SDFA of a deterministic SDAG defines the stochastic semantics of the SDAG. 
In turn, the stochastic semantics of the SDAG of an SDFA is specified by the stochastic language of the SDFA.

In a DFG discovered by a conventional discovery technique, every action has only a single corresponding node. 
Every SDAG with a pair of distinct nodes that refer to the same action $\lambda$ can be transformed to an SDAG in which those two nodes are removed, a fresh node for $\lambda$ is added, all the incoming (outgoing) arcs of the removed nodes get rerouted to reach (originate at) the fresh node, and the probabilities on the outgoing arcs of the fresh node are normalized.
Repeated application of this
transformation until no further reductions are feasible yields an
SDAG that is a DFG.\footnote{Noting
that in a DFG nodes and arcs are annotated with occurrence frequencies.}
It is straightforward to show that different maximal sequences of feasible transformations do indeed lead to the same resulting DFG.
\Cref{fig:SDAG:2} shows the DFG $G'$ obtained in this way from SDAG $G$ in \Cref{fig:SDAG:1}; we denote this relationship between the graphs by $G'=\funcCall{DFG}{G}$.

An SDAG can be annotated with frequencies of actions and flows, thereby providing information on the rate at which the corresponding concepts arise in the event log from which the SDAG was constructed.
Given a number $n$ of cases that should ``flow'' through the graph, frequencies can be derived from the probabilities by solving a system of equations comprising two types.
Each node of the graph defines a \emph{conservation} equation that requires that the sum of frequencies on the incoming arcs equal the sum of the frequencies on the outgoing arcs, except for the input (output) node, for which the sum of frequencies on the outgoing (incoming) arcs is equal to $n$.
Finally, each arc emanating from a node $s$ defines an \emph{arc} equation that specifies that the frequency of the arc is equal to its probability times the sum of frequencies of all the incoming arcs of $s$.
Such a system of equations always has a solution as it contains one unknown per arc and at least as many equations as arcs.

The annotations in \Cref{fig:SDAG:1,fig:SDAG:2} show frequencies for nodes and arcs, obtained from the probabilities by following that process.
For instance, the annotations in \Cref{fig:SDAG:2} were obtained by solving the fourteen simultaneous equations in {\Cref{table:system:equations}}.\footnote{In which $\funcCallTwo{f}{x}{y}$ is the frequency of arc $\pair{x}{y} \in \gamma$.}
Notice that in the figures, all frequencies are rounded to one decimal place.
The frequency of a node is derived as the sum of the frequencies of all its incoming (or outgoing) arcs.

\begin{table}[!tp]
\vspace{-5mm}
\centering
\caption{\small A system of equations used to obtain frequencies of nodes and arcs in \Cref{fig:SDAG:2}.}
\vspace{1mm}
\hspace{-2mm}
\scriptsize
\begin{minipage}{.45\textwidth}
\begin{tabular}{c|r@{}l} %
\toprule 
\,\, \textbf{Node}  \,\, & \multicolumn{2}{c}{\,\, \textbf{Equation}  \,\,} \\
\midrule
  &  &  \\
$i$	& $1493.0$ & $\;=\,\funcCallTwo{f}{i}{n_{1}}$ \\
$n_1$ & $\funcCallTwo{f}{i}{n_1}$ & $\;=\,\funcCallTwo{f}{n_1}{n_2} + \funcCallTwo{f}{n_1}{n_3}$ \\
$n_2$ & $\funcCallTwo{f}{n_1}{n_2}$ & $\;=\,\funcCallTwo{f}{n_2}{n_3}$ \\
$n_3$ & $\funcCallTwo{f}{n_1}{n_3} + \funcCallTwo{f}{n_2}{n_3} + \funcCallTwo{f}{n_5}{n_3}$ & $\;=\,\funcCallTwo{f}{n_3}{o} + \funcCallTwo{f}{n_3}{n_5}$ \\
$n_5$ & $\funcCallTwo{f}{n_3}{n_5}$ & $\;=\,\funcCallTwo{f}{n_5}{n_3} + \funcCallTwo{f}{n_5}{o}$ \\
$o$ & $\funcCallTwo{f}{n_3}{o} + \funcCallTwo{f}{n_5}{o}$ & $\;=\,1493.0$ \\
 & & \\
\bottomrule
\end{tabular}
\end{minipage}
\hspace{8mm}
\begin{minipage}{.45\textwidth}
\begin{tabular}{c|r@{\;=\;}l} %
\toprule 
\,\, \textbf{Arc}  \,\, & \multicolumn{2}{c}{\,\, \textbf{Equation}  \,\,} \\
\midrule
$\funcCallTwo{\gamma}{n_1}{n_2}$ & $0.17 \; \funcCallTwo{f}{i}{n_1}$ & $\funcCallTwo{f}{n_1}{n_2}$ \\
$\funcCallTwo{\gamma}{n_1}{n_3}$ & $0.83 \; \funcCallTwo{f}{i}{n_1}$ & $\funcCallTwo{f}{n_1}{n_3}$ \\
$\funcCallTwo{\gamma}{n_2}{n_2}$ & $0.17 \; (\funcCallTwo{f}{n_1}{n_2} + \funcCallTwo{f}{n_2}{n_2})$ & $\funcCallTwo{f}{n_2}{n_2}$ \\
$\funcCallTwo{\gamma}{n_2}{n_3}$ & $0.83 \; (\funcCallTwo{f}{n_1}{n_2} + \funcCallTwo{f}{n_2}{n_2})$ & $\funcCallTwo{f}{n_2}{n_3}$ \\
$\funcCallTwo{\gamma}{n_3}{n_5}$ & $0.50 \; (\funcCallTwo{f}{n_1}{n_3} + \funcCallTwo{f}{n_2}{n_3} + \funcCallTwo{f}{n_5}{n_3})$ & $\funcCallTwo{f}{n_3}{n_5}$ \\
$\funcCallTwo{\gamma}{n_3}{o}$ & $0.50 \; (\funcCallTwo{f}{n_1}{n_3} + \funcCallTwo{f}{n_2}{n_3} + \funcCallTwo{f}{n_5}{n_3})$ & $\funcCallTwo{f}{n_3}{o}$ \\
$\funcCallTwo{\gamma}{n_5}{n_3}$ & $0.80 \; \funcCallTwo{f}{n_3}{n_5}$ & $\funcCallTwo{f}{n_5}{n_3}$ \\
$\funcCallTwo{\gamma}{n_5}{o}$ & $0.20 \; \funcCallTwo{f}{n_3}{n_5}$ & $\funcCallTwo{f}{n_5}{o}$ \\
\bottomrule
\end{tabular}
\end{minipage}
\label{table:system:equations}
\vspace{-5mm}
\end{table}

The SDAG in {\Cref{fig:SDAG:1}} is of size 16 and has an
{\method{entropic relevance}} of $3.267$ bits per trace relative to
the example event log $L$ from \Cref{sec:preliminaries}, with the
small difference with the relevance of the SDFA in {\Cref{fig:SDFA}}
due to the integer frequencies used in the SDAG.
The DFG in {\Cref{fig:DAG:DFvM}} constructed from the same log using
the {\DFvM} algorithm~\cite{Leemans2019a} has size 19 and an
{\method{entropic relevance}} to $L$ of $4.168$.
That is, the SDAG is smaller and also describes the event log more
faithfully than the DFG.
Nor does varying the {\DFvM} filtering threshold in steps of $0.01$
from zero to $1.0$ find any outcomes that alter that relativity: 70
of those 100 generated DFGs have a size of 10 and relevance of
$6.062$; 19 DFGs are of size 14 and relevance of $4.804$; and the
remaining 11 DFGs (as in {\Cref{fig:DAG:DFvM}}) have size of 19 and
relevance of $4.168$.
Finally, the relevance of the SDAG in {\Cref{fig:SDAG:2}} to $L$ is
$4.865$ bits per trace, illustrating the desirability of permitting
duplicate action nodes in the discovered models.
\vspace{-2mm}
\section{Stochastic Process Discovery}
\label{sec:spd}
\vspace{-1mm}

Our approach to discovering stochastic process models consists of two
fundamental components: a {\emph{grammar inference}} algorithm
coupled with a {\emph{genetic optimization}} mechanism.
We build on {\ALERGIA} to construct representations of stochastic
language models encoded in the given event data.
Complementing that model discovery, we employ
{\method{Multi-Objective Genetic Search}} to fine-tune the parameters
associated with the learning process.
This section explains these two fundamental components.

{\ALERGIA}, an instantiation of the {\method{Red-Blue}} algorithm, is introduced by Carrasco and Oncina~{\cite{Carrasco1994}} for learning SDFAs from a multiset of strings.
The algorithm starts by constructing a {\emph{Prefix Acceptor Tree}} (PAT) from the multiset of traces, with nodes representing prefixes of traces and edges indicating transitions between the nodes.
Each edge is labeled with the frequency of the corresponding trace prefix in the input log.
{\ALERGIA} then generalizes and compacts that PAT by merging states that exhibit similar behavior.
The merging aims to identify sets of states with similar probabilistic distributions of outgoing edges, and consolidates each such set into a single state, thereby seeking to create a more concise representation of the underlying language of the model.

Two subsets of nodes are maintained while the compaction process is carried out.
The {\Red} set initially contains only the root of the prefix tree, while the {\Blue} set contains all direct successors of the {\Red} set.
At each cycle of operation, {\ALERGIA} iteratively selects a state from the {\Blue} set, taking into account a threshold parameter {\(t\)} that determines the minimum number of strings necessary for a state to be considered for merging.
Compatibility for merging is determined by comparing final-state frequencies and outgoing transitions of states in the {\Red} and {\Blue} sets.

{\emph{Hoeffding's inequality}}~{\cite{H63JASA}} is a statistical test that bounds the extent to which the mean of a set of observations can deviate from its expected value.
This inequality can be used to test if an observed outcome $g$ significantly deviates from a probability value $p$ given a confidence level $\alpha$ and $n$ observations:
\begin{equation}\label{eq:hoeffding:inequality}
	\left| p-\frac{g}{n} \right| <
		\sqrt{\frac{1}{2n} \log \frac{2}{\alpha}} \, .
\end{equation}
{\ALERGIA} approximates that inequality to compare the observed frequencies of pairs of states drawn from the {\Red} and {\Blue} sets, taking into account final-state frequencies and the frequencies of outgoing transitions:
\begin{equation}\label{eq:hoeffding}
	\left| \frac{g_1}{n_1} - \frac{g_2}{n_2} \right|
		< \omega \cdot \left(\sqrt{\frac{1}{n_1}}
		+ \sqrt{\frac{1}{n_2}}\,\right) \, .
\end{equation}
In~\Cref{eq:hoeffding}, $n_1$ and $n_2$ are the frequencies of arriving sequences at the compared states, and $g_1$ and $g_2$ are the frequencies of ending sequences at those states.
This Hoeffding bound assesses the significance of the difference between the two estimates.
It ensures that the difference in frequencies is within a statistically acceptable range, as determined by the value of $\omega$.
If the observed frequencies fall within the permitted range, the states are deemed compatible, and {\ALERGIA} merges them.
Conversely, if the observed differences exceed the bounds, the states are considered incompatible, and the algorithm preserves that difference by converting the {\Blue} member of the pair to {\Red}.
These compatibility checks are not limited to the states but also extend to their respective successors in the tree structure.
To merge states, the algorithm redirects transitions and folds subtrees rooted at an identified {\Blue} state onto the outgoing edges of the corresponding {\Red} state.
Based on the merged transitions, the automaton is then updated, with the process iterated until no more mergings can be identified.

\begin{figure}[t]
\centering
\begin{minipage}{0.8\textwidth}
\begin{algorithm}[H]
\footnotesize
  \KwData{A multiset of traces $\eventLog{}$, parameters $\omega > 0$, $t > 0$, and $f \in \intervalcc{0}{1}$}
  \KwResult{An SDFA $\automaton{}$}
  $\eventLog{}' \leftarrow \var{FILTER}(\eventLog{}, f)$\;\label{algo:fun:filter}
  $T \leftarrow \var{PAT}(\eventLog{}')$\;\label{algo:fun:pat}
  $\Red \leftarrow \{q_0\}$\;
  $\Blue \leftarrow \{\delta(q_0, a) : \forall a \in \alphabet\}$\;
  \While{$\exists q_b \in \Blue : \var{FREQ}(q_b) \geq t$}{\label{algo:fun:freq}
    \If{$\exists q_r \in \Red : \var{COMPATIBLE}(q_r, q_b, \omega)$}{\label{algo:fun:compatible}
      $\automaton{} \leftarrow \var{MERGE}(q_r, q_b)$\;\label{algo:fun:merge}
      $\automaton{} \leftarrow \var{FOLD}(q_r, q_b)$\;\label{algo:fun:fold}
    }
    \Else{
      $\Red \leftarrow \Red \cup \{q_b\}$\;
      $\Blue \leftarrow (\Blue - \{q_b\}) \cup\{\delta(q_b, a) : a \in \alphabet \mbox{ {\bf{and}} } \delta(q_b, a) \not\in \Red\}$\;
    }
  }
  $\automaton{} \leftarrow \var{CONVERT}(T)$\; \label{algo:fun:convert}
\KwRet{$\automaton{}$}\;
\caption{\ALERGIA}
\label{algo:alergia}
\end{algorithm}
\end{minipage}
\begin{minipage}{0.8\textwidth}
\begin{algorithm}[H]
\footnotesize
\KwData{Initial population size $n$, number of generations $\var{GenLim}$, and number of parents $k$ to generate offspring}
\KwResult{A Pareto frontier $F$, captured as a set of parameter triples}
$g \leftarrow 0$\;
$P \leftarrow \var{POPULATION}(n)$\;\label{algo:fun:population}
\While{$g < \var{GenLim}$}{
    $F \leftarrow \var{SELECT}(P)$\;\label{algo:fun:select}
    $U \leftarrow \var{CROSSOVER-MUTATION}(F,k)$\;\label{algo:fun:crossover}
    $P \leftarrow \var{REPLACE-ELITE}(U,F)$\;\label{algo:fun:replace}
    $g \leftarrow g + 1$\;
}
\KwRet $F$\;
\caption{\GASPD}
\label{algo:moga}
\end{algorithm}
\vspace{-8mm}
\end{minipage}
\end{figure}

As a pre-processing step, we employ a simple filtering technique that selectively retains a certain percentage, as a predefined threshold $f$, of the most frequent traces from the event log.
This filtering reduces the complexity of the discovered model, particularly in scenarios where noise and infrequent traces might detract from the overall view of the process.
Varying the filtering threshold $f$ also allows altering the level of detail preserved in the log, allowing alignment with specific analytical goals.

{\Cref{algo:alergia}} provides an overview of {\ALERGIA}. 
The $\var{FILTER}$ function, called at {\cref{algo:fun:filter}}, filters the input log based on the supplied threshold.
The filtered log is then used to construct PAT $T$, refer to {\cref{algo:fun:pat}}. 
{\Cref{fig:PAT}} shows the prefix tree constructed from example log $\eventLog{}$ from {\Cref{sec:preliminaries}} after applying the filtering using the threshold of $f=0.89$. 
At~{\cref{algo:fun:freq}}, the $\var{FREQ}$ function computes the frequency of arriving at state $q_b$ to check if the state is reached sufficiently frequently, as per the threshold $t$.
The compatibility test between states $q_r$ and $q_b$ is performed by the $\var{COMPATIBLE}$ function at {\cref{algo:fun:compatible}}.
The $\var{MERGE}$ function is then called at~{{\cref{algo:fun:merge}}} to merge two states and, subsequently, the $\var{FOLD}$ function folds the tree.
Finally, the $\var{CONVERT}$ function maps $T$ to its corresponding automaton $\automaton{}$.
For a detailed description of the algorithm, refer to~{\cite{Carrasco1994}}.
The SDFA $\automaton{}$ in~{\Cref{fig:SDFA}} is discovered from the example event log by {\ALERGIA} using parameters $\omega = 1$, $t = 1$, and $f = 0.89$.

The initialization phase sets the foundation for the genetic algorithm.
At \cref{{algo:fun:population}} of {\Cref{algo:moga}}, the $\var{POPULATION}$ function creates a population of $n$ seed solutions $P = \set{p_1, p_2, \ldots, p_n}$, where $p_i = \triple{\omega_{i}}{{t}_{i}}{f_{i}}$, $i \in \intintervalcc{1}{n}$, is a parameter triple.
The values of these parameters are independently and randomly generated within specified bounds, with each triple determining a model using {\Cref{algo:alergia}}.
A large initial population enhances exploration but increases the computational cost.
Conversely, a small initial population is computationally more efficient but may compromise the ability to find good solutions.

The quality of each solution $p_i$ is then assessed to determine how well it performs with respect to the objective functions.
In {\Cref{algo:moga}}, this is done in function $\var{SELECT}$ at \cref{{algo:fun:select}}, which applies {\ALERGIA} to obtain a process model as a function of each parameter triple in the current population~$P$.
This gives rise to a set of models, where each model has an associated size and an {\method{entropic relevance}} score computed using {\ENTROPIA}~\cite{Polyvyanyy2020a}.
The selection of relevance as a quality metric stems from its ability to rapidly score models, a feature that harmonizes effectively with the genetic framework.

As part of each selection phase, the individuals from the current population that form the {\emph{Pareto frontier}} $F$ are identified, noting the individuals that are not dominated by other solutions in terms of the two objectives, with Pareto efficient points leading to models having small {\method{size}} and {\method{entropic relevance}}.
Each such point represents a unique trade-off between the objectives, providing decision-makers with a set of alternative options.
One such frontier is associated with each {\emph{generation}}, with the generations counted in {\Cref{algo:moga}} by the variable $g$.
While we use {\ALERGIA}, {\method{size}}, and {\method{entropic relevance}}, the {\GASPD} procedure is not tied to a specific discovery algorithm or quality measure.

In function $\var{CROSSOVER-MUTATION}$ at \cref{{algo:fun:crossover}} of {\Cref{algo:moga}}, each generation is constructed from the previous one by applying crossover and mutation operations to create a set of offspring, new individuals that (with luck) inherit beneficial traits from previous generations.
The {\emph{crossover}} operation mimics natural genetic recombination, combining information from selected parents, with both single- and double-point crossover techniques employed in our approach.
Specifically, to produce offspring, two parents $\triple{\omega_1}{{t}_{1}}{f_1}$ and $\triple{\omega_2}{{t}_{2}}{f_2}$ are selected from frontier $F$. 
In the single-point crossover, one crossover position in the parent triples is selected.
Then, the parameters at and to the left of the crossover point in both parents remain unchanged, while the parameters to the right of the crossover point are swapped between the parents.
Hence, six offspring are produced, namely 
$\triple{\omega_1}{t_2}{f_2}$ and $\triple{\omega_2}{t_1}{f_1}$ at position one, $\triple{\omega_1}{t_1}{f_2}$ and $\triple{\omega_2}{t_2}{f_1}$ at position two, and $\triple{\omega_1}{{t}_{1}}{f_1}$ and $\triple{\omega_2}{{t}_{2}}{f_2}$ at position three.
In the double-point crossover, two additional offspring $\triple{\omega_1}{t_2}{f_1}$ and $\triple{\omega_2}{t_1}{f_2}$ are produced by swapping parameters of the parents twice, once after the first crossover point and then after the second crossover point.
{\GASPD} includes all such offspring stemming from all pairs in $k$ randomly selected parents from front $F$ (or $\cardinality{F}$ selected parents, if $\cardinality{F} < k$) when generating the set of offspring $U$.

The {\emph{mutation}} component of the $\var{CROSSOVER-MUTATION}$ function then alters the ``genetic makeup'' of new individuals in $U$ by randomly modifying their three defining parameters, restricted to certain predefined bounds.
The random nature of these mutations ensures that the algorithm does not converge prematurely to a limited subset of solutions and allows for the exploration of a broader range of possibilities, crucial when the global optimum may not be immediately apparent.
For each individual $\triple{\omega}{t}{f}$ in $U$, its mutated twin defined as $\triple{\omega + \Delta_{\omega}}{t + \Delta_{t}}{f + \Delta_{f}}$, where $\Delta_{\omega}$, $\Delta_{t}$, and $\Delta_{f}$ are random positive or negative mutation values within specified bounds, is created and added to $U$.

Our approach here (implemented in the $\var{REPLACE-ELITE}$ function at \cref{algo:fun:replace} of {\Cref{algo:moga}}) is that each generation retains the items that were on the Pareto frontier of any previous generation, and then adds any new parameter triples that establish new points on the frontier.
That is, we always preserve solutions that, at some stage, have appeared promising, and seek to add any new solutions that outperform them.

The SDFAs constructed by {\ALERGIA} using the parameters discovered by {\GASPD} can be translated to sound SDAGs using the principles laid out in \Cref{def:SDFA2SDAG}.
\section{Evaluation}
\label{sec:evaluation}

We have implemented {\GASPD}{\footnote{\url{https://github.com/jbpt/codebase/tree/master/jbpt-pm/gaspd}}}
and conducted experiments using twelve publicly available real-world
event logs shared by the IEEE Task Force on Process
Mining{\footnote{\url{https://www.tf-pm.org/resources/logs}}},
derived from IT systems executing business processes.
These experiments assess the feasibility of using {\GASPD} in
industrial settings, and compare the quality of its models
with the ones constructed by {\DFvM}~{\cite{Leemans2019a}}.
A wide range of {\DFvM} filtering parameters were considered,
establishing a reference curve for each log showing the versatility
of {\DFvM} across the spectrum of possible model sizes.
We also explored the effectiveness of the genetic search in guiding
the selection of parameters to identify desirable solutions.

For each of the twelve event logs {\GASPD} was seeded with an initial
population of $50$ parameter triples, each containing random values
for $\omega$ (from $0$ to $15$); for $t$ (from $0$ to the most
frequent PAT branch); and for $f$ (from $0$ to $1$).
The genetic search was then iterated for $50$ generations, with
mutation achieved by random adjustments to parameter values within the
same defined ranges.

\begin{figure*}[t]
\vspace{-3mm}
\centering
\subfloat[ Generation 1 ]{\includegraphics[clip=true, trim=0cm 0cm 0cm 0cm, scale=0.22]{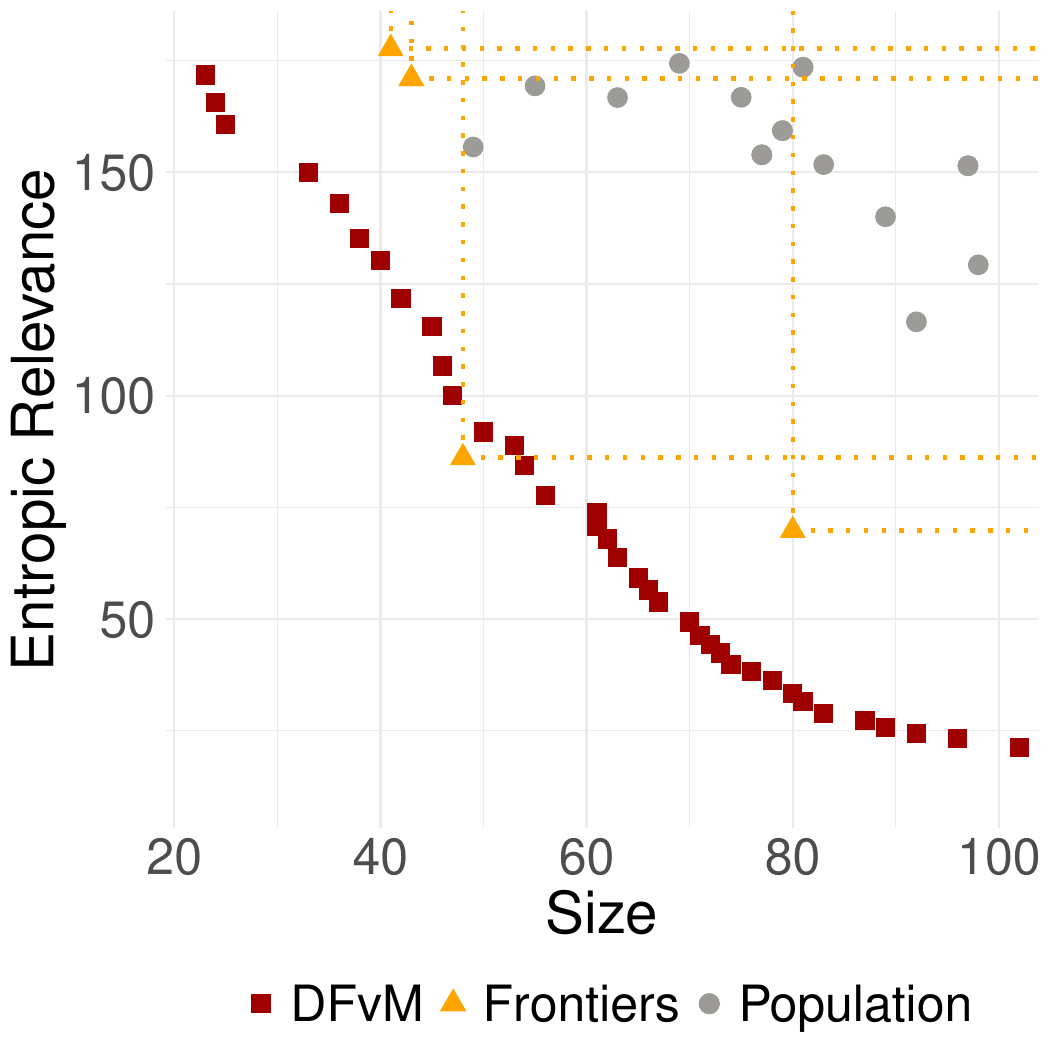}
\label{fig:plot:bpic17:g1}}%
\hfill
\subfloat[ Generation 10 ]{\includegraphics[clip=true, trim=0cm 0cm 0cm 0cm, scale=0.22]{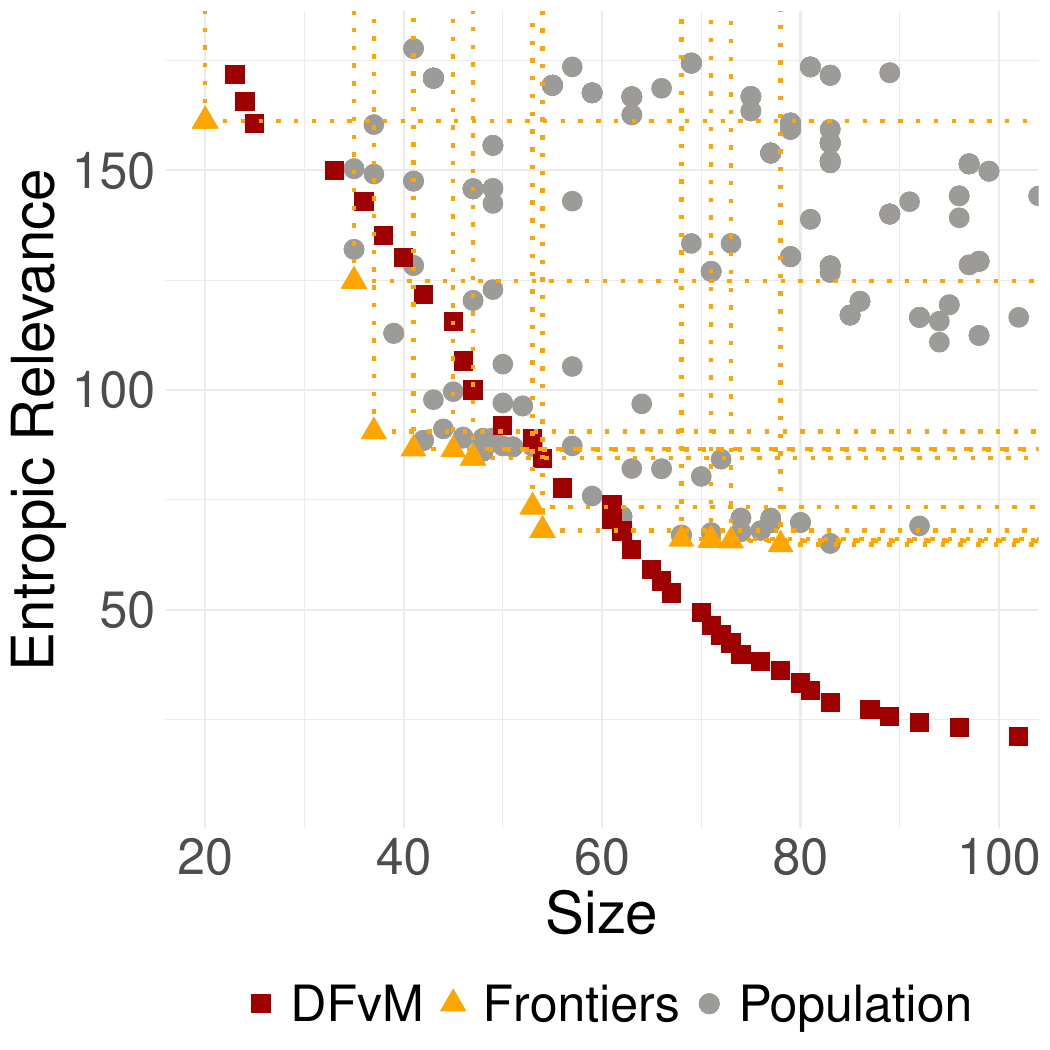}
\label{fig:plot:bpic17:g10}}%
\hfill
\subfloat[ Generation 50 ]{\includegraphics[clip=true, trim=0cm 0cm 0cm 0cm, scale=0.22]{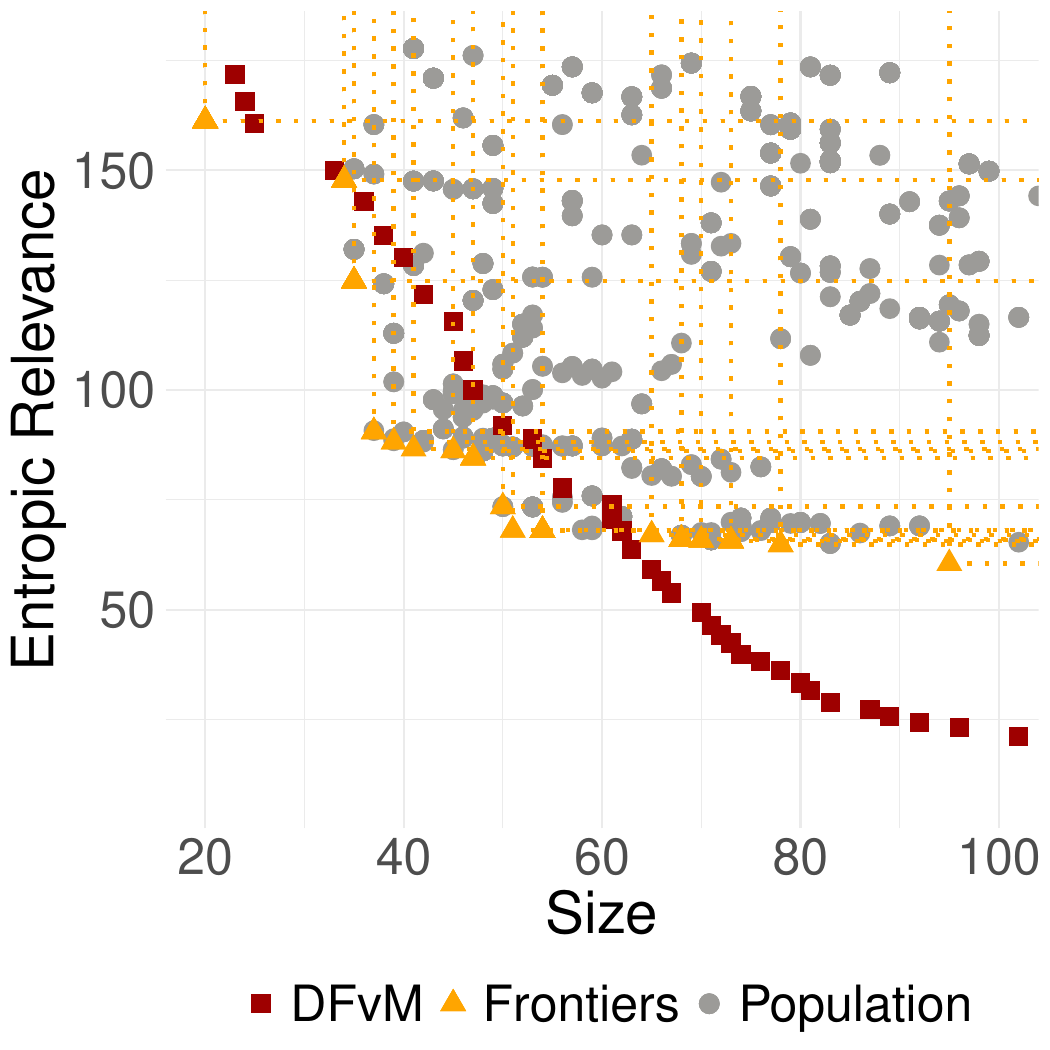}
\label{fig:plot:bpic17:g50}}\\
\vspace{-1mm}
\caption{\small Size and relevance of SDAG models discovered by {\GASPD}
from the BPIC17 log.}
\label{fig:plots:generations}%
\vspace{-3mm}
\end{figure*}

{\Cref{fig:plot:bpic17:g1,fig:plot:bpic17:g10,fig:plot:bpic17:g50}}
show model relevance scores as a function of model size,
taking three snapshots during the course of {\GASPD} when
executed on the BPIC17 log.
After one generation the models are a mixed bag, a result
of the random starting point; but with broad coverage of the search
space achieved, and already with individuals (yellow dots) identified
that outperform the baseline set by the red {\DFvM} frontier.
By the tenth generation, mutation and breeding have taken the
population toward improved performance, with many solutions now below
the previous frontier, and convergence towards promising regions in
the parameter space.
Then, by generation $50$, the situation has stabilized, with
additional individuals identified that lead to attractive models that
outperform the {\DFvM} frontier used here as a reference.

\begin{figure*}[!tp]
\vspace{-1mm}
\centering
\subfloat[Sepsis]{\includegraphics[clip=true, trim=0cm 0cm 0cm 0cm, scale=0.22]{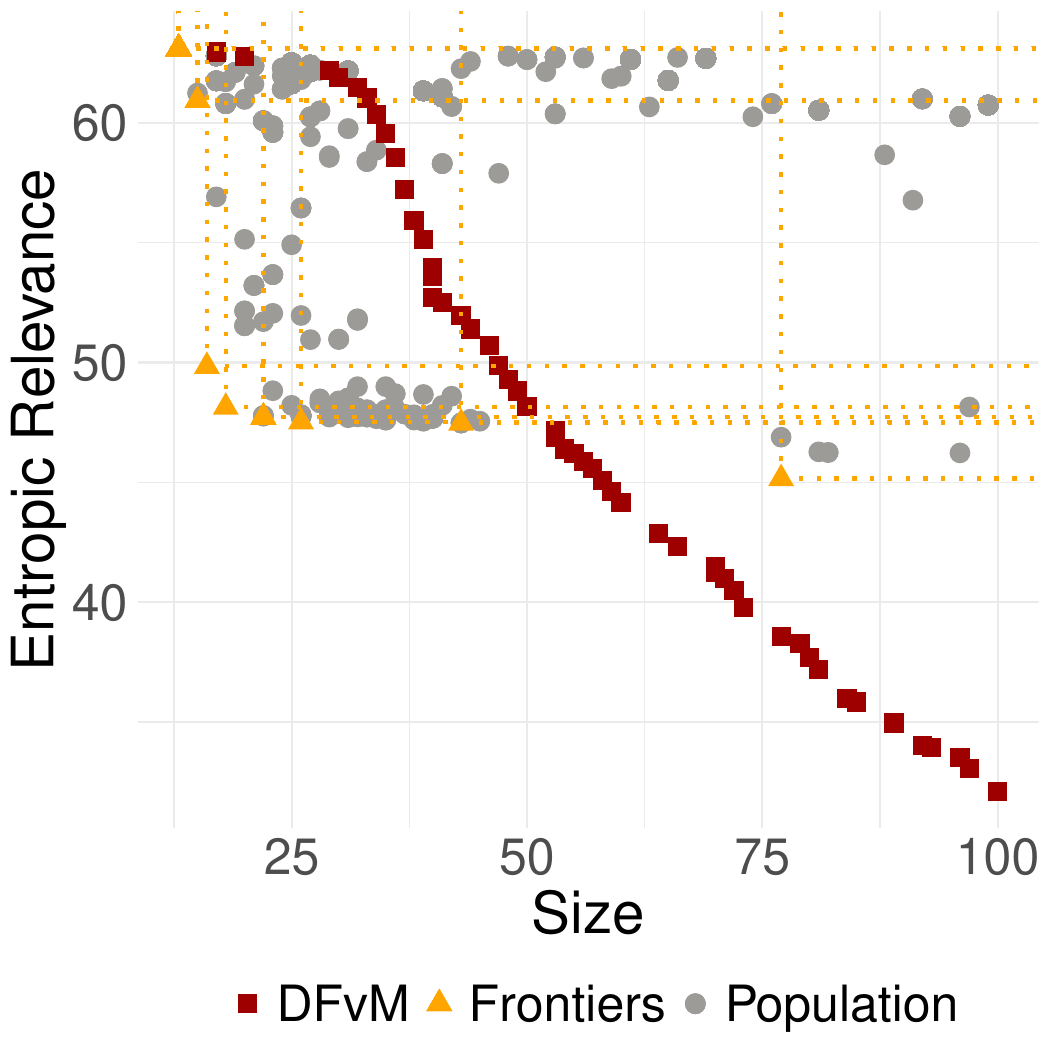}
\label{fig:plot:sepsis:g50}}
\hfill
\subfloat[BPIC12]{\includegraphics[clip=true, trim=0cm 0cm 0cm 0cm, scale=0.22]{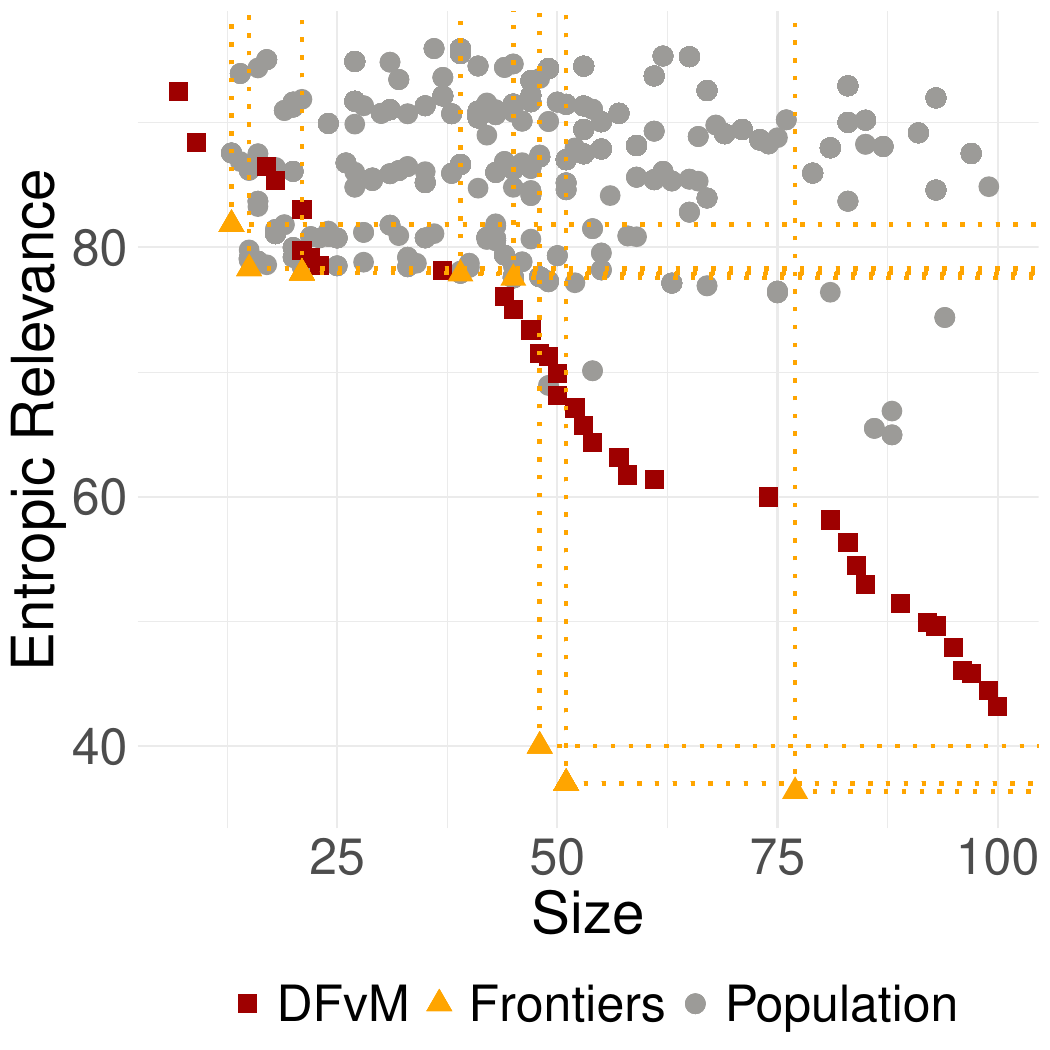}
\label{fig:plot:bpic12:g50}}
\hfill
\subfloat[BPIC13I]{\includegraphics[clip=true, trim=0cm 0cm 0cm 0cm, scale=0.22]{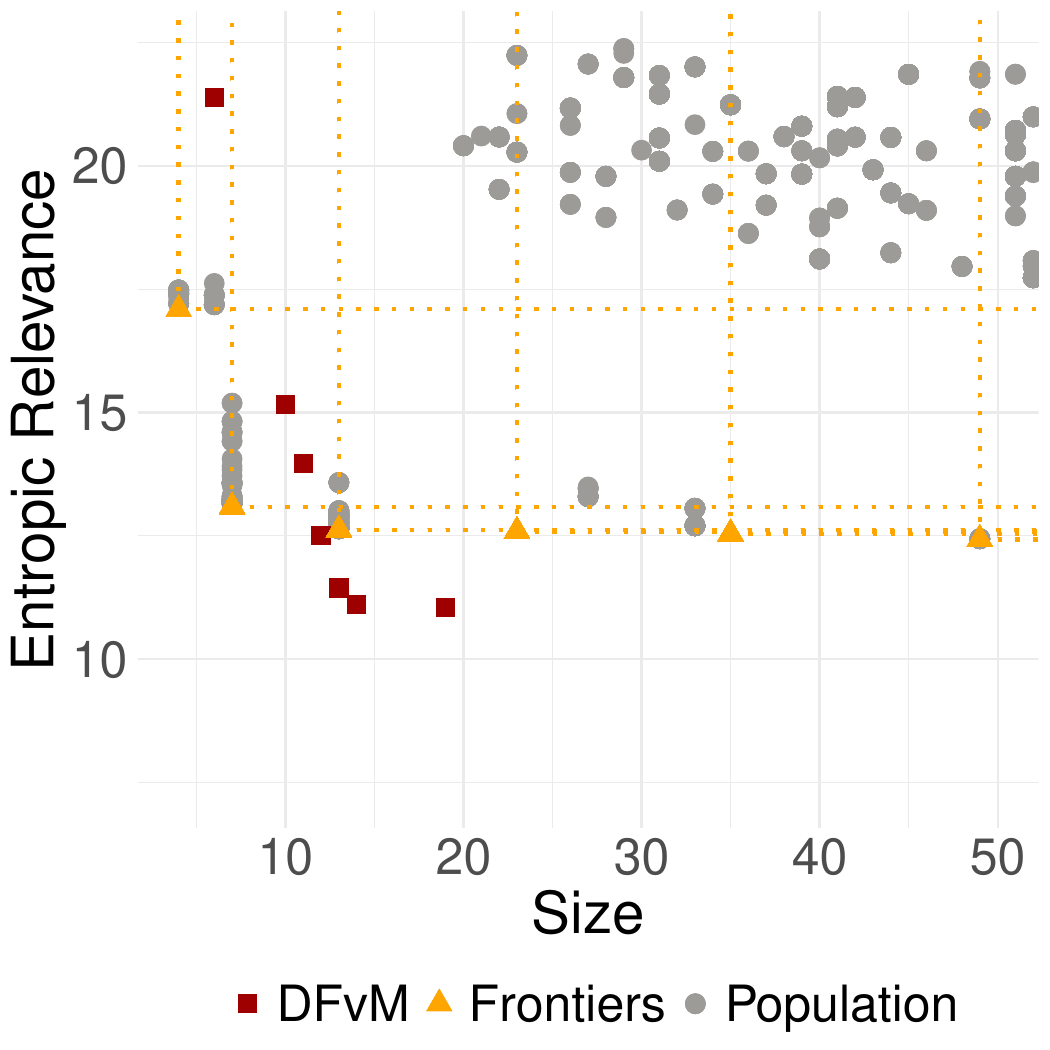}
\label{fig:plot:bpic13i:g50}}
\vspace{-1mm}
\caption{\small Size and relevance of SDAG models discovered by {\GASPD}
after 50 generations.}
\label{fig:plots:results}%
\vspace{-2mm}
\end{figure*}

{\Cref{fig:plots:results}} shows {\GASPD} performance at the $50$
generation point for three other logs.
In each case {\GASPD} constructs models that outperform the {\DFvM}
ones over at least some fraction of the range of sizes being
considered, noting that smaller and more accurate models are
preferable for human analysis.
{\Cref{table:intervals}} then summarizes {\GASPD} when applied to all
twelve event logs, further supporting our contention that {\GASPD}
finds useful new models.
In the first seven rows we focus on human-scale models of up to $100$
nodes/edges, and then relax that to a limit of size $1000$ for the
BPIC15 tasks, which tend to give rise to more complex models.
As can be seen, in ten of the twelve cases {\GASPD} generates models
of better relevance for at least part of the size range, with {\DFvM}
tending to discover models of better relevance for large sizes.
In future work, it would be interesting to study if grammar inference
techniques can be useful in discovering large, accurate models.

As already described, {\GASPD} employs evolutionary search as part of
the discovery algorithm.
To focus the search onto good candidates, the individuals in each
generation $P$ are split into two classes: ``good'', and ``bad''.
An individual is ``good'' if it has been on the Pareto frontier in any
previous generation; and it is ``bad'' if has never been part of any
Pareto frontier.
In {\Cref{algo:moga}} each generation is derived solely from the
``good'' individuals of the previous population, with the aim of
iterating towards better solutions.
Only new individuals that are also ``good'' are then retained into
the next population.

\begin{table}[!tp]
\vspace{-3mm}
\centering
\caption{\small Relative performance of SDAG models discovered by {\GASPD} after $50$
generations, including the size range of interest, the size interval(s)
of {\GASPD} superiority, min/max size, and min/max relevance.}
\vspace{1mm}
\scriptsize
\begin{tabular}{l|rr|c|rr|rr} %
    \toprule
    & \multicolumn{2}{c|}{\textbf{\,\,\,Size interval\,\,\,}} 
    & \multicolumn{1}{c|}{\textbf{\,\,\,Size interval of\,\,\,}} 
    & \multicolumn{2}{c|}{\textbf{\,Size within interval\,}} 
    & \multicolumn{2}{c}{\textbf{\,\,\,Entropic relevance\,\,\,}} \\
    \multirow{-1}{*}{\textbf{\begin{tabular}[c]{@{}c@{}}\,\,\,\,\,Event log\,\,\,\,\,\end{tabular}}} & 
    \multirow{-1}{*}{\textbf{\begin{tabular}[c]{@{}c@{}}\,\,From\,\,\end{tabular}}} & 
    \multirow{-1}{*}{\textbf{\begin{tabular}[c]{@{}c@{}}\,\,\,\,\,\,To\,\,\,\,\,\,\end{tabular}}} &
    \multirow{-1}{*}{\textbf{\begin{tabular}[c]{@{}c@{}}\,\,\,\,\,superior performance\,\,\,\,\,\end{tabular}}} &
    \multirow{-1}{*}{\textbf{\begin{tabular}[c]{@{}c@{}}\,\,\,\,\,\,Min\,\,\,\,\,\,\end{tabular}}} &
    \multirow{-1}{*}{\textbf{\begin{tabular}[c]{@{}c@{}}\,\,\,\,\,\,Max\,\,\,\,\,\,\end{tabular}}} & 
    \multirow{-1}{*}{\textbf{\begin{tabular}[c]{@{}c@{}}\,\,\,\,\,\,Min\,\,\,\,\,\,\end{tabular}}} &
    \multirow{-1}{*}{\textbf{\begin{tabular}[c]{@{}c@{}}\,\,\,\,\,\,Max\,\,\,\,\,\,\end{tabular}}} \\
    \midrule
    \begin{filecontents*}{data/table.csv}
log;from;to;interval;minsize;maxsize;minrel;maxrel
Sepsis;0;100;$\intintervaloo{0}{53}$;13;77;45.16;63.10
BPIC12;0;100;$\intintervalco{13}{44} \cup \intintervalcc{48}{100}$;13;77;36.39;81.84
BPIC17;0;100;$\intintervaloo{0}{25} \cup \intintervalco{34}{62}$;20;95;60.56;161.25
RTFM;0;100;$\intintervalcc{53}{100}$;13;69;2.71;9.07
BPIC13OP;0;100;$\intintervalco{4}{6} \cup \intintervalco{8}{9}$;3;80;5.06;7.80
BPIC13CP;0;100;$\intintervaloo{0}{9}$;4;79;6.26;9.78
BPIC13I;0;100;$\intintervaloo{0}{12}$;4;89;12.33;17.10
BPIC15-1;0;1000;$\intintervalco{37}{44} \cup \intintervalco{59}{122}$;37;633;380.01;384.21
BPIC15-2;0;1000;$\intintervalco{473}{641} \cup \intintervalco{697}{965}$;149;946;428.00;469.94
BPIC15-3;0;1000;$\intintervaloo{0}{27} \cup \intintervalco{95}{103}$;11;958;298.01;370.63
BPIC15-4;0;1000;$\emptyset$;475;475;386.93;386.93
BPIC15-5;0;1000;$\emptyset$;695;695;445.41;445.41
\end{filecontents*}
\csvreader[head to column names, late after line=\\, separator=semicolon]{data/table.csv}{} %
    {\,\,\,\,\log & \from\,\,\,\,\,\,\,\, & \to\,\,\,\,\,\, & \interval & \minsize\,\,\,\,\,\, & \maxsize\,\,\,\,\,\, & \minrel\,\,\,\, & \maxrel\,\,\,\, } %
    \bottomrule
\end{tabular}
\label{table:intervals}
\vspace{-5mm}
\end{table}

\begin{figure*}[th]
\centering
\includegraphics[clip=true, trim=0cm 0cm 0cm 0cm, scale=0.49]{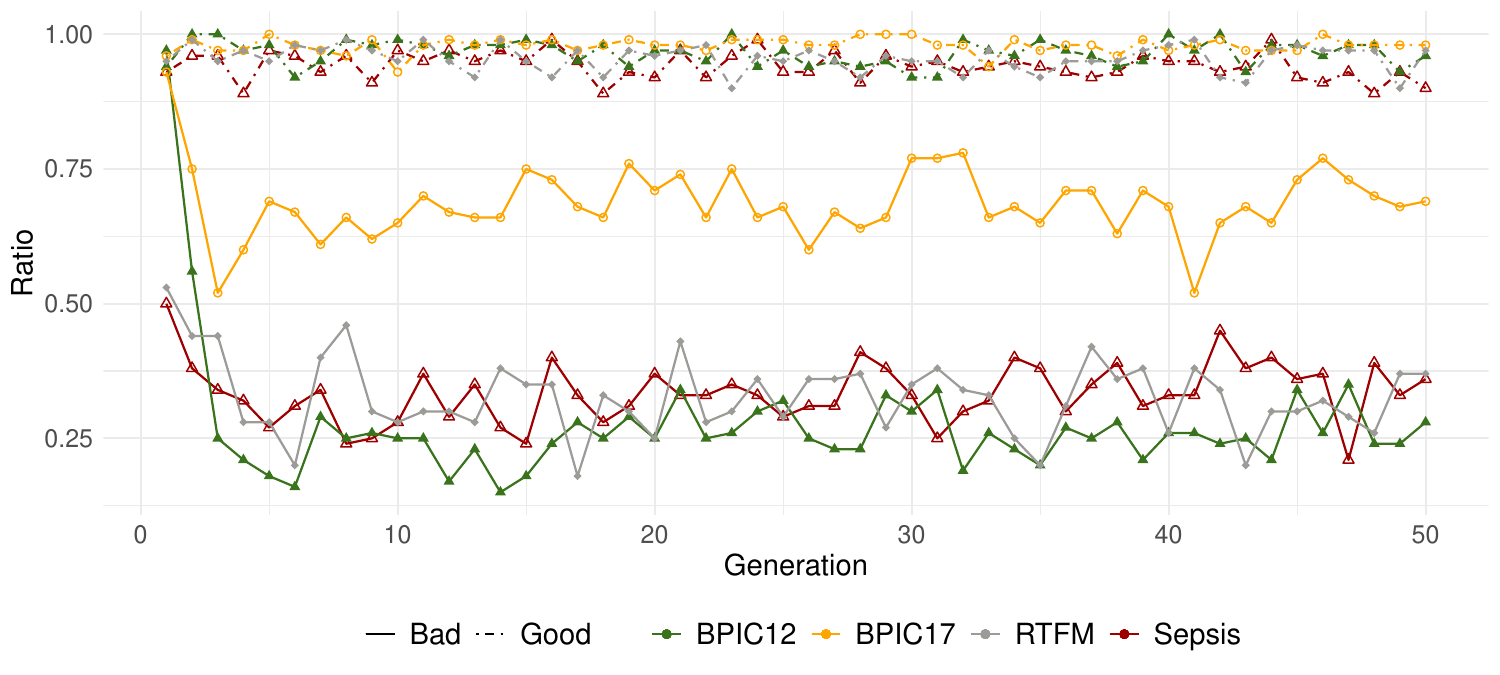}
\vspace{-3mm}
\caption{\small The fraction of ``good'' individuals arising as
offspring from pairs of ``good'' parents and pairs of ``bad''
parents, traced through $50$ generations, for each of four logs.
}
\vspace{-3mm}
\label{fig:plot:ratios}%
\end{figure*}

To verify the usefulness of that heuristic, we also experimented with
breeding from ``bad'' parents, recording at each generation the
fractions of ``good'' offspring from pairs of ``bad'' parents, and
then likewise the fraction arising from pairs of ``good'' parents.
{\Cref{fig:plot:ratios}} shows the result, with the four dotted lines
at the top the ``good parents'' success rate, and the four solid
lower lines the ``bad parents'' rate.
As can be seen, the four dotted lines comprehensively outperform
their solid equivalents, and good parents are much more likely to
lead to interesting offspring than are bad parents.
Taking a paired $t$-test between the $50$ ``dotted'' points and the
$50$ ``solid'' points for the four logs gives $p<10^{-10}$ in all
four cases, confirming that filtering the population at each
generation to only contain good parents is a highly beneficial
strategy.
\section{Discussion and Conclusion}
\label{sec:discussion:conclusion}

We have presented {\GASPD}, an algorithm for discovering sound Stochastic
Directed Action Graphs, a variant of the DFGs often used in commercial process mining applications.
The technique is grounded in grammar inference over sequences of
actions, as recorded in event logs; and uses a bespoke genetic
algorithm that ensures fast convergence towards models of practically
interesting sizes and accuracy.

{\GASPD} can be used to implement a user-friendly tool for exploring
input logs, with
interesting models often uncovered even in the first generation of
random parameters, and then improved through subsequent generations.
Moreover, construction of multiple such models is easily done in
parallel, meaning that as the interesting models become progressively
available, they can be presented to the user via a dynamic slider
interface~\cite{Polyvyanyy2008b}, ordered by size.
This conjectured interface can thus present the so-far discovered
Pareto-best models, while generations are extended and new models are
constructed in the background, further feeding the slider.
A second slider can be introduced to navigate over all the computed
generations.
For instance, selecting a specific generation in this slider can load
all the Pareto optimal models obtained after that generation in the
other slider.
Such controls would support interactive exploration of the
improvement in model quality observed through generations.

The exploration of alternative grammatical inference techniques
stands out as a promising avenue for enhancing the quality of the
discovered models.
Replacing {\ALERGIA} with other grammar inference methods in the
genetic framework can reveal their ultimate effectiveness in
generating process models from event logs.
{\ALERGIA} appears as a strong candidate for initiating the quest due
to its acknowledged performance characteristics over a wide variety
of real-world languages~\cite{Higuera2010}.
Additionally, there is a prospect for improvement by introducing
pruning strategies over prefix acceptor trees before merging states
and redesigning the merging rules.
These enhancements can streamline the algorithm's convergence
process, ensuring more efficient solution space exploration,
resulting in more accurate and concise process models and allowing
the discovery of superior models of larger sizes.
Finally, new ideas to guide the genetic exploration of the parameter
space of grammar inference can be explored.
Some initial ideas include using simulated annealing and particle
swarm optimization principles to escape parameter subspaces of local
optimal models.

The evaluation in {\Cref{sec:evaluation}} made use of entropic
relevance and model size to identify Pareto-optimal models.
Entropic relevance ensures models that better describe the
frequencies of log traces are prioritized, and strikes a balance
between conventional precision and recall quality measures in process
mining~\cite{AlkhammashPMG22IS}; and model size is a standard measure
of simplicity for DFGs~\cite{A19CENTERIS}.
But other measurement combinations could also be considered, and
would lead to different concrete measurements and, consequently,
might result in different Pareto-optimal models.
Nevertheless it seems probable that measures that assess the same
broad phenomena as entropic relevance and size will result in similar
conclusions to those achieved here -- that {\GASPD} provides the
ability to realize interesting and useful process models not found by
other current approaches, making it an important development for
process mining.\vspace{-1.5mm}
\bibliography{strings.bib,bib.bib}
\newpage
\end{document}